%% file: main.tex
\def\year{2022}\relax
\documentclass[letterpaper]{article}
\usepackage{aaai22}  
\usepackage{times}  
\usepackage{helvet}  
\usepackage{courier}  
\usepackage[hyphens]{url}  
\usepackage{graphicx} 
\usepackage{natbib}  
\usepackage{caption} 
\DeclareCaptionStyle{ruled}{labelfont=normalfont,labelsep=colon,strut=off} 
\usepackage{algorithm}
\usepackage{algorithmic}

\usepackage{newfloat}
\usepackage{listings}
\floatstyle{ruled}
\newfloat{listing}{tb}{lst}{}
\floatname{listing}{Listing}

\usepackage{amsmath,amsfonts,bm}
\usepackage{multirow, adjustbox, booktabs, colortbl}
\usepackage{subcaption} 
\usepackage{color}

\newtheorem{definition}{Definition}
\usepackage{ulem}

\title{Pointspectrum: Equivariance Meets Laplacian Filtering for Graph Representation Learning}
\author {
    Marinos Poiitis,\textsuperscript{\rm 1}
    Pavlos Sermpezis, \textsuperscript{\rm 1}
    Athena Vakali \textsuperscript{\rm 1}
}
\affiliations {
    \textsuperscript{\rm 1} Aristotle University of Thessaloniki\\
    mpoiitis@csd.auth.gr, sermpezis@csd.auth.gr, avakali@csd.auth.gr
}

\begin{document}

\maketitle

\begin{abstract}
Graph Representation Learning (GRL) has become essential for modern graph data mining and learning tasks. GRL aims to capture the graph's structural information and exploit it in combination with node and edge attributes to compute low-dimensional representations. While Graph Neural Networks (GNNs) have been used in state-of-the-art GRL architectures, they have been shown to suffer from over smoothing when many GNN layers need to be stacked. In a different GRL approach, \textit{spectral methods} based on graph filtering have emerged addressing over smoothing; however, up to now, they employ traditional neural networks that cannot efficiently exploit the structure of graph data. Motivated by this, we propose PointSpectrum, a spectral method that incorporates a set equivariant network to account for a graph's structure. PointSpectrum enhances the efficiency and expressiveness of spectral methods, while it outperforms or competes with state-of-the-art GRL methods. Overall, PointSpectrum addresses over smoothing by employing a graph filter and captures a graph's structure through set equivariance, lying on the intersection of GNNs and spectral methods. Our findings are promising for the benefits and applicability of this architectural shift for spectral methods and GRL.
%
\end{abstract}

\section{Introduction}

Graphs are universal mathematical structures --usually accompanied by a plethora of features-- that are extensively used to describe real-world data 
in various domains such as citation and social networks~\cite{kipf2016semi, liu2019single}, recommender systems~\cite{zhang2019inductive}, or adversarial attacks~\cite{zhang2020gnnguard}. Due to the complex structure of graphs, traditional machine learning models are insufficient for addressing graph-based tasks such as node and graph classification, link prediction and node clustering. This necessity has given rise to \textit{Graph Representation Learning (GRL)} methods, which aim to capture the structure of input graph data and produce meaningful low-dimensional representations.

A main track of GRL methods is based on Graph Neural Networks (GNNs). GNNs and specifically Graph Convolutional Networks (GCN)~\cite{kipf2016semi} have advanced the research on GRL leading to outstanding performance as they capture a node's neighborhood influence. Nevertheless, each GNN layer considers only the local one-hop node relations leading to unavoidably deep layer stacking to efficiently account for the global graph information. Notwithstanding, vanilla GNNs cannot be arbitrarily deep as they lead to over smoothing and information loss ~\cite{xu2018representation, zhao2019pairnorm, chen2020measuring}. To address over smoothing, spectral methods exploiting graph filters have been introduced~\cite{wang2019attributed, zhang2019attributed}. However, so far, these methods have been used in conjunction with traditional neural networks (e.g., MLPs or CNNs) that cannot efficiently exploit the \textit{set equivariance} property of graph data (i.e., equivariance to permutations in the input data). In the graph domain there is no implicit data ordering, and thus the existing spectral methods waste an important portion of their computational capacity. 

\noindent\textbf{Contribution.} In this work, we propose \textit{PointSpectrum}, a GRL architecture that bridges the gap between GNN-based and spectral GRL methods (Section~\ref{sec:methodology}). PointSpectrum is based on Laplacian smoothing (used in spectral methods to alleviate the problem of over-smoothing), while it maintains the set equivariance property of GNN-based approaches.

Specifically, PointSpectrum is an unsupervised end-to-end trainable architecture, consisting of the following components:
\begin{itemize}
    \item \textit{Input}: a low-pass graph filter (Laplacian smoothing) is applied on the input graph data that enables the computation of k-order graph convolution for arbitrarily large k without over smoothing node features
    \item \textit{Encoder}: the input data are fed to a set equivariant network that generates low-dimensional node embeddings 
    \item \textit{Decoder}: a joint loss is employed to account for the reconstruction of the input data and their better separation in the embedding space through a clustering metric.
\end{itemize}

\begin{figure*}[t]
    \centering
    \includegraphics[width=0.7\linewidth]{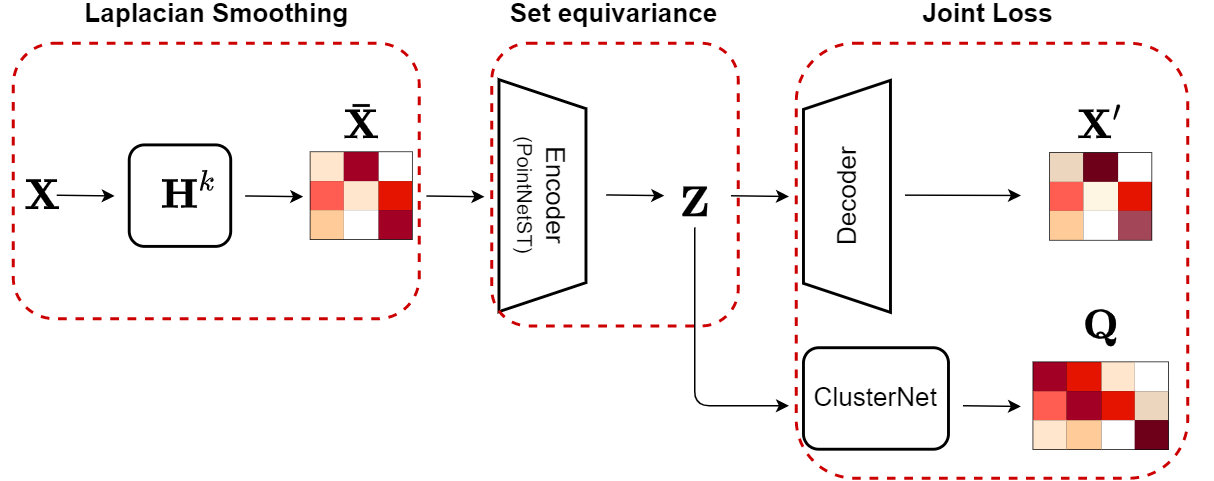}
    \caption{PointSpectrum architecture. Encoder is a pointNetST network producing node embeddings in an equivariant manner, the pair-wise decoder reconstructs input $\mathbf{\bar{X}}$ and last ClusterNet is employed to better separate points in the embedding space.}
    \label{fig:pointSpectrum_arch}
\end{figure*}

To the best of our knowledge this is the first work that introduces set equivariance in spectral GRL methods. Our experimental results (Section~\ref{sec:results}) show that: (i) Using set equivariant networks can increase the robustness (wrt. the model parameters) and efficiency (e.g., faster convergence, expressiveness) of spectral methods. (ii) PointSpectrum presents high performance in all benchmark tasks and datasets, outperforming or competing with the state-of-the-art (detailed in Section~\ref{sec:related}). 
Overall, our findings showcase a new direction for GRL: the combination of spectral methods with set equivariance. Incorporating set equivariant networks in existing spectral methods can be straightforward (e.g., replacing MLPs or CNNs), and our results are promising for the efficiency of this approach.

\section{Methodology}
\label{sec:methodology}

\noindent \textbf{Definitions.} We consider a non-directed attributed graph $\mathcal{G} = (\mathcal{V}, \mathcal{E}, \mathbf{X})$, where $\mathcal{V} = \{v_1, v_2, ..., v_n\}$ is the node set with $|\mathcal{V}| = n$, $\mathcal{E}$ is the edge set, and $\mathbf{X} = [\mathbf{x}_1, \mathbf{x}_2, \dots, \mathbf{x}_n]^T \in \mathbb{R}^{n \times l}$ is the feature matrix consisting of feature vectors $\mathbf{x}_i \in \mathbb{R}^l, ~ \forall v_i \in \mathcal{V}$. The structural information of graph $\mathcal{G}$ is represented by the adjacency matrix $\mathbf{A} = \{a_{ij}\} \in \mathbb R_{\ge 0}^{n \times n}$.

\noindent \textbf{Goals.} The goal of GRL is to map nodes to low-dimensional embeddings, which should preserve both the structural ($\mathbf{A}$) and contextual ($\mathbf{X}$) information of $\mathcal{G}$. We denote as $\mathbf{Z} \in \mathbb{R}^{n \times l'}$, with $l' \ll l$, the matrix with the node embeddings.

\noindent \textbf{Approach overview.} We propose a methodology to compute an embedding matrix $\mathbf{Z}$, which combines structural and contextual information in a twofold way: on one hand, it uses Laplacian smoothing (based on $\mathbf{A}$) of the feature matrix $\mathbf{X}$ (Section~\ref{sec:smoothing}) and, on the other hand, by considering the node features as a set of points it exploits global information using a permutation equivariant network architecture (Section~\ref{sec:equivariance}). Last, we leverage node clustering as a boosting component that further enhances performance by better separating nodes in the embedding space. The entire architecture that brings these components together is presented in Section~\ref{sec:PointSpectrum}.

\subsection{Graph Convolution and Laplacian Smoothing}
\label{sec:smoothing}

The most important notion in the prevalent GNN-based embedding methods, such as GCN~\cite{kipf2016semi}, is that neighboring nodes should be similar and hence their features should be smoother than non-neighboring nodes in the graph manifold. However, these methods capture deeper connections by stacking multiple layers, leading to deep architectures, which are known to overly smooth the node features~\cite{chen2020measuring}. Over smoothing occurs as each layer repeatedly smooths the original features so as to account for the deeper interactions. To address this problem, the domain of graph signal processing has been used and in particular graph convolution by using Laplacian Smoothing filters~\cite{zhang2019attributed, cui2020adaptive}. 

Specifically, ``spectral methods'' in GRL use a smoothed feature matrix $\mathbf{\Bar{X}}$, instead of the original $\mathbf{X}$, which corresponds to a k-th order graph convolution of $\mathbf{X}$:
\begin{equation}\label{eq:X-bar-k-definition}
    \mathbf{\Bar{X}} = \mathbf{H^k}\mathbf{X}
\end{equation}
where matrix $\mathbf{H} \in \mathbb{R}^{n \times n}$ is the Laplacian Smoothing filter (discussed below). The multiplication of feature matrix $\mathbf{X}$ with filter $\mathbf{H}$ corresponds to a 1-order graph convolution (or 1-order graph smoothing). Stacking $k$ filters together, i.e., $k$ multiplications with $\mathbf{H}$, leads to a $k$-order convolution as in Eq.~\ref{eq:X-bar-k-definition}. Thus, deep network interactions are captured by the power of filter $\mathbf{H}$ instead of repeated convolutions of the input features and therefore over smoothing is avoided.

\noindent \textbf{The Laplacian Smoothing Filter, $\mathbf{H}$:} The intuition behind k-order convolution of spectral methods is the following: Each column of feature matrix $\mathbf{X(:,i)} \in \mathbb{R}^n$ (i.e., the vector with the values of all nodes for a single feature) can be considered as a graph signal. The smoothness of a signal depicts the similarity between the graph nodes. Since neighboring nodes should be similar, to capture this similarity, we would need to construct a smooth signal based on $\mathbf{X(:,i)}$ that captures node adjacency and takes into account the features of the most important neighboring nodes. It can be shown (details in Appendix~\ref{sec:smoothing-full}) that the Laplacian Smoothing filter $\mathbf{H}$ as defined in Eq.~\ref{eq:generalized_laplacian}~\cite{taubin1995signal} can cancel high frequencies between neighboring nodes and preserve the low ones:
\begin{equation}
\label{eq:generalized_laplacian}
    \mathbf{H} = \mathbf{I} - \mu\mathbf{L} 
\end{equation}
where $\mu \in \mathbb{R}$, $\mathbf{I}$ is the identity matrix and $\mathbf{L}$ is the graph Laplacian. The graph Laplacian is defined as $\mathbf{L} = \mathbf{D} - \mathbf{A}$, where $\mathbf{D} = diag(d_1, d_2, \dots, d_n) \in \mathbb{R}^{n \times n}$ is the degree matrix of $\mathbf{A}$, with $d_i = \sum_{v_j \in \mathcal{V}}a_{ij}$ being the degree of node $v_i$. In order for $\mathbf{H}$ to be low-pass, $1-\mu\lambda$ should be a non-negative degressive function. \cite{cui2020adaptive} showed that the optimal value of $\mu$ is $1/\lambda_{max}$, with $\lambda_{max}$ denoting the largest eigenvalue of $\mathbf{L}$; in the remainder, we use this value for $\mu$.

\begin{figure*}[t]
	\begin{subfigure}[t]{0.32\linewidth}
		\includegraphics[width=1\linewidth]{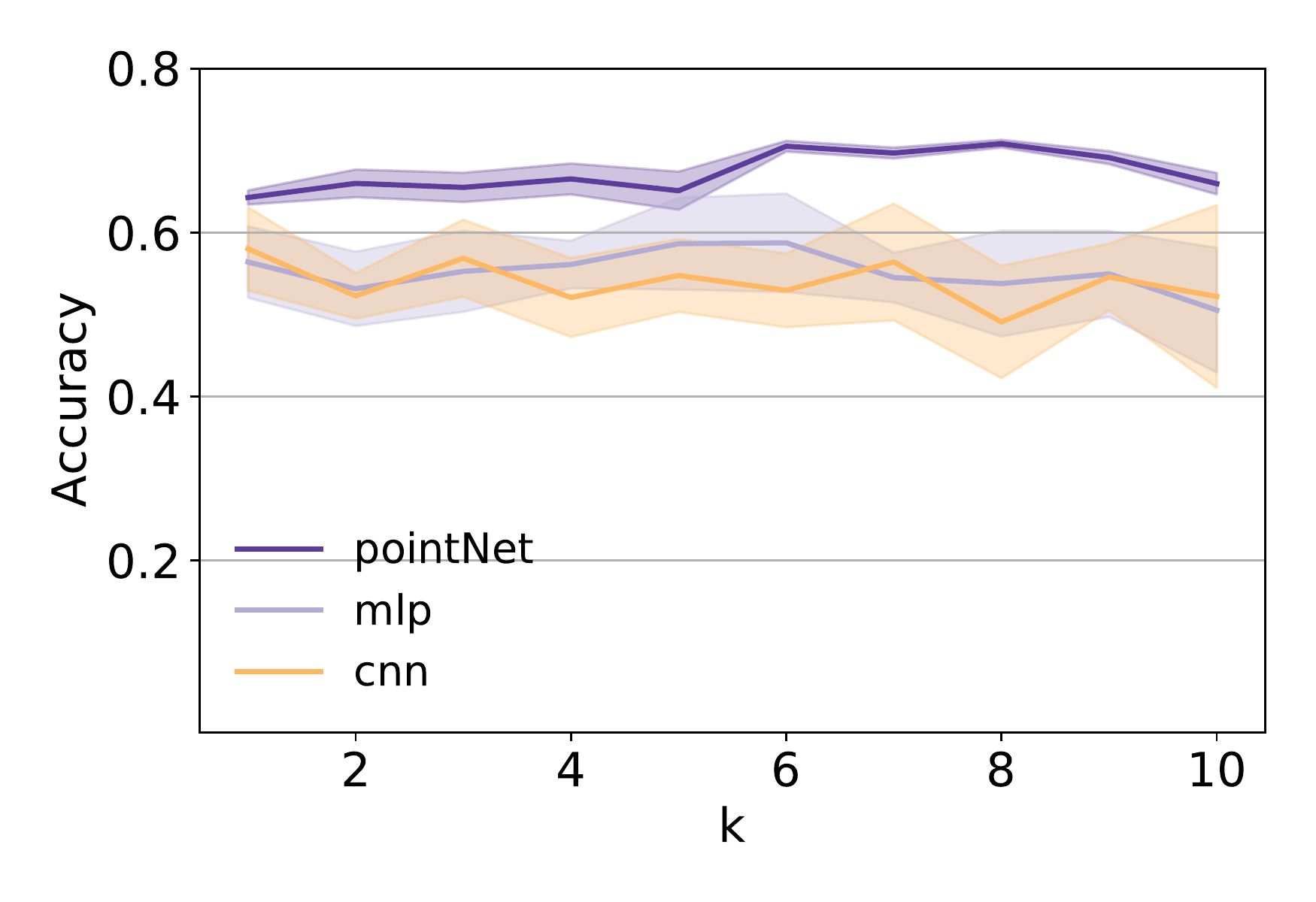}
    		\caption{Accuracy (Cora)}\label{fig:cora_acc}		
	\end{subfigure}
    \hfill
	\begin{subfigure}[t]{0.32\linewidth}
		\includegraphics[width=1\linewidth]{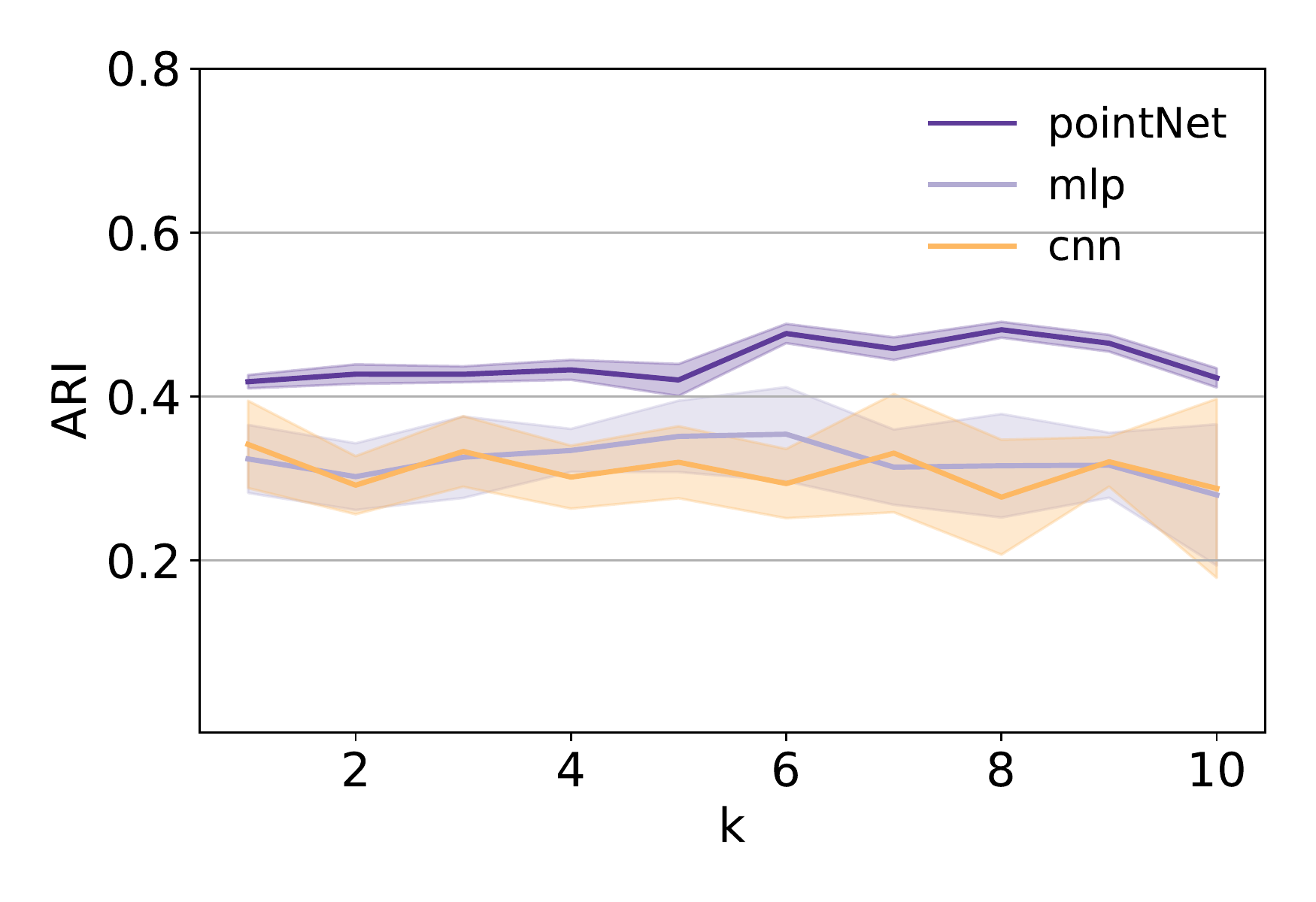}
		\caption{ARI (Cora)}\label{fig:cora_ari}	
	\end{subfigure}
    \hfill
	\begin{subfigure}[t]{0.32\linewidth}
		\includegraphics[width=1\linewidth]{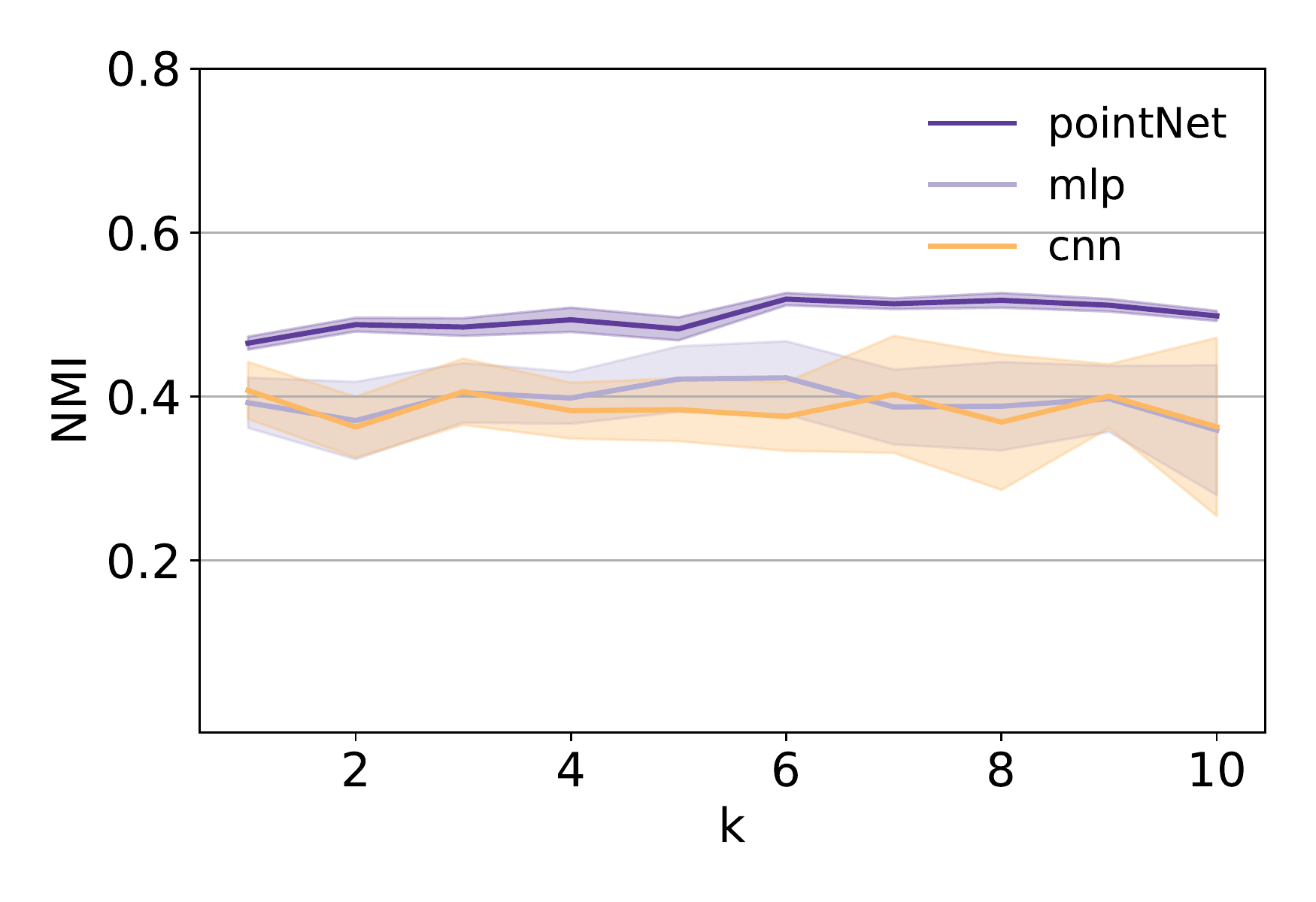}
		\caption{NMI (Cora)}\label{fig:cora_nmi}		
	\end{subfigure}
	\caption{Accuracy, ARI and NMI metrics of the PointSpectrum with PointNetST, MLP and CNN-based networks in the encoder solving the clustering task on the Cora dataset. For all metrics, the mean value and standard deviation of 10 experiment runs are depicted. \textit{The set equivariant PointNetST achieves higher performance and is more robust than the MLP and CNN variants, irrespective of the convolution order.}}
	\label{fig:different_k_s}
\end{figure*}

\noindent\textbf{Renormalization trick:} In practice the renormalization trick~\cite{kipf2016semi}, i.e., adding self-loops in the graph, has been shown to improve accuracy and shrink the graph spectral domain~\cite{wu2019simplifying}. Qualitatively, this means that for every node the smoothing filter also considers its own features alongside the ones from its neighbors. Thus, we perform the following transformation: self-loops are added to the adjacency: $\mathbf{\bar{A}} = \mathbf{I} + \mathbf{A}$; we then use the symmetric normalized graph Laplacian $\mathbf{\bar{L}_{sym}} = \mathbf{\bar{D}^{-\frac{1}{2}}}\mathbf{\bar{L}}\mathbf{\bar{D}^{-\frac{1}{2}}}$, where $\mathbf{\bar{D}}$ and $\mathbf{\bar{L}}$ are the degree and Laplacian matrices of $\mathbf{\bar{A}}$. Finally, the resulting Laplacian smoothing filter becomes $\mathbf{H} = \mathbf{I} - k\mathbf{\bar{L}_{sym}}$.

\subsection{Equivariance and PointNetST}\label{sec:equivariance}

Neural network architectures can approximate any continuous function $f$ given sufficient capacity and expressive power~\cite{sonoda2017neural, sannai2019universal, yarotsky2021universal}; here, we denote a neural network as a function $f: \mathbf{X} \mapsto Y$ operating on the feature matrix $\mathbf{X}$ (and by extension on $\mathbf{\bar{X}}$).

\noindent\textbf{Sets and permutation equivariance.} Conventional neural networks such as Multilayer Perceptrons (MLPs) or Convolutional Neural Networks (CNNs) act on data where there is an implicit order (e.g., adjacent pixels in images). However, graphs do not have any implicit order: nodes can be presented in different order but the graph still maintains the same structure (isomorphism). To this end, $\mathbf{X}$ can be seen as a set of points $\mathbf{x} \in \mathbb{R}^l$, and $f$ as a function operating on a set. 

\begin{definition}
\label{prop:equivariance}
A function $f: \mathbb{R}^{n \times l} \mapsto \mathbb{R}^{n \times l'}$ acting on a set $\mathbf{X} \in \mathbb{R}^{n \times l}$ is \emph{permutation equivariant} when $f(\sigma \cdot \mathbf{X}) = \sigma \cdot f(\mathbf{X})$ for any permutation $\sigma$ of the set $\mathbf{X}$.
\end{definition}
In other words, a \textit{permutation equivariant} function $f$ produces the same output (e.g., embedding, label) for each set item (e.g., node) irrespective of the order of the given data. This means that if a transformation (e.g., a permutation of the input matrix's rows) is applied to the input data, then the same transformation should be applied to the model's output as well. 

\noindent\textbf{Equivariance in GRL methods.}
Set equivariance is not captured by traditional neural networks such as MLPs (or CNNs), which are mainly used in the ``spectral methods'' in GRL~\cite{zhang2019attributed, cui2020adaptive}. More specifically, \cite{zaheer2017deep} proved that a function $f$ is permutation equivariant iff it can be decomposed in the form $\rho (\sum_{\mathbf{x} \in \mathbf{X}} \phi (\mathbf{x}))$, where $\rho$ and $\phi$ are suitable transformations
. Also, each MLP or CNN layer can be seen as a transformation $\phi$, and a deep network with $m$ such layers can be expressed as $f(\mathbf{X}) = \phi_m \circ \dots \circ \phi_1(\mathbf{X})$. Hence, the necessary decomposition for permutation equivariance does not hold. On the contrary, GNNs can capture not only permutation equivariance but also graph connectivity, which makes them even more expressive. Nevertheless, as mentioned earlier, this high expressiveness comes with oversmoothing problems. \textit{In this work, we aim to bring the benefits of both approaches used in GRL: we employ a set equivariant network that accounts for the graph structure (through the matrix $\mathbf{\bar{X}}$) and avoids oversmoothing at the same time (as discussed in Section~\ref{sec:smoothing})}. This property is crucial for this work's model prevalence over the traditional neural networks, as it will be shown in Section~\ref{sec:results}. 

\begin{figure*}[t]
	\begin{subfigure}[t]{0.32\linewidth}
		\includegraphics[trim=0mm 7mm 0mm 0mm,clip,width=1.05\linewidth]{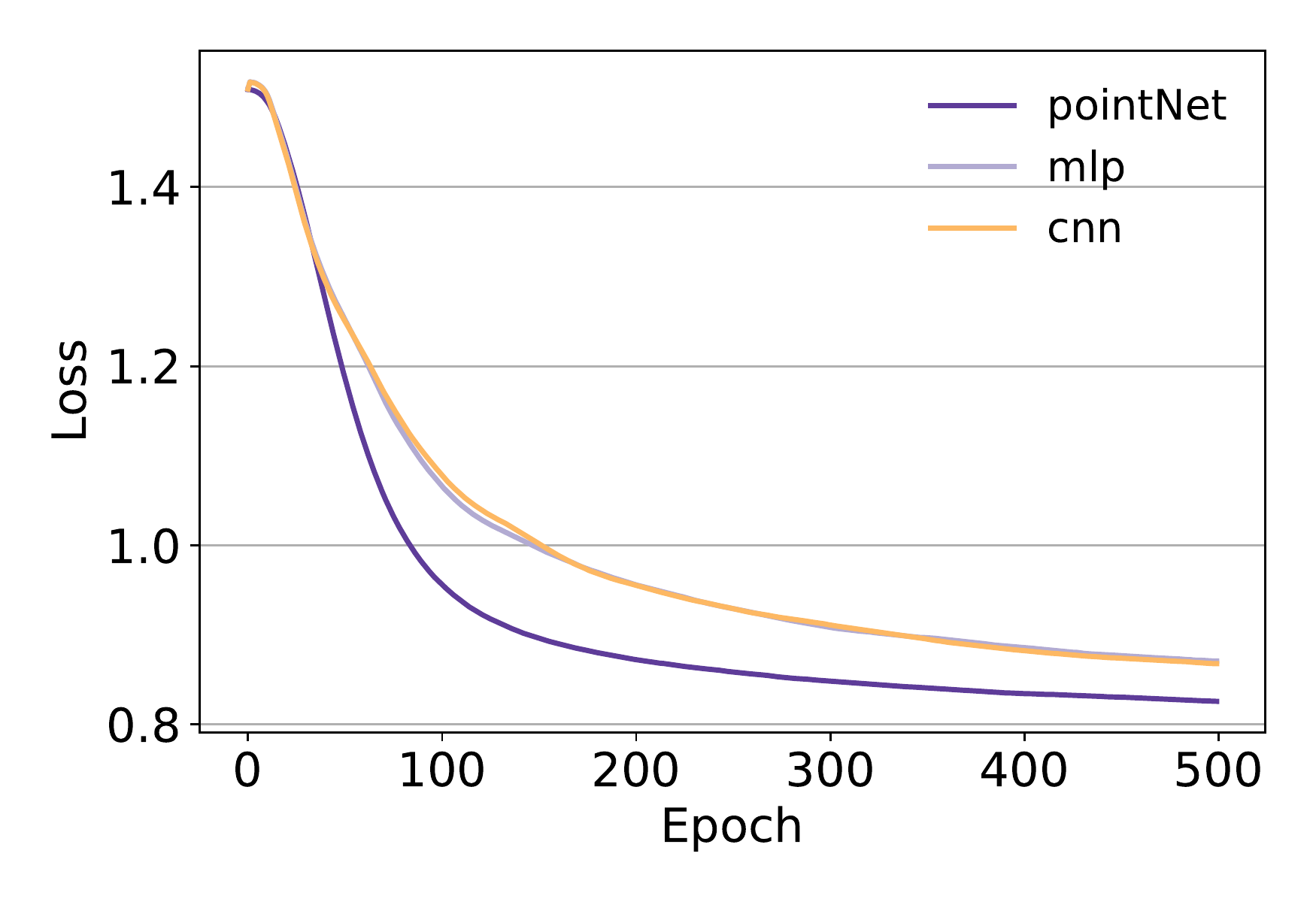}
    		\caption{Loss (Cora)}\label{fig:cora_loss}		
	\end{subfigure}
    \hfill
	\begin{subfigure}[t]{0.32\linewidth}
		\includegraphics[trim=0mm 7mm 0mm 0mm,clip,width=1.05\linewidth]{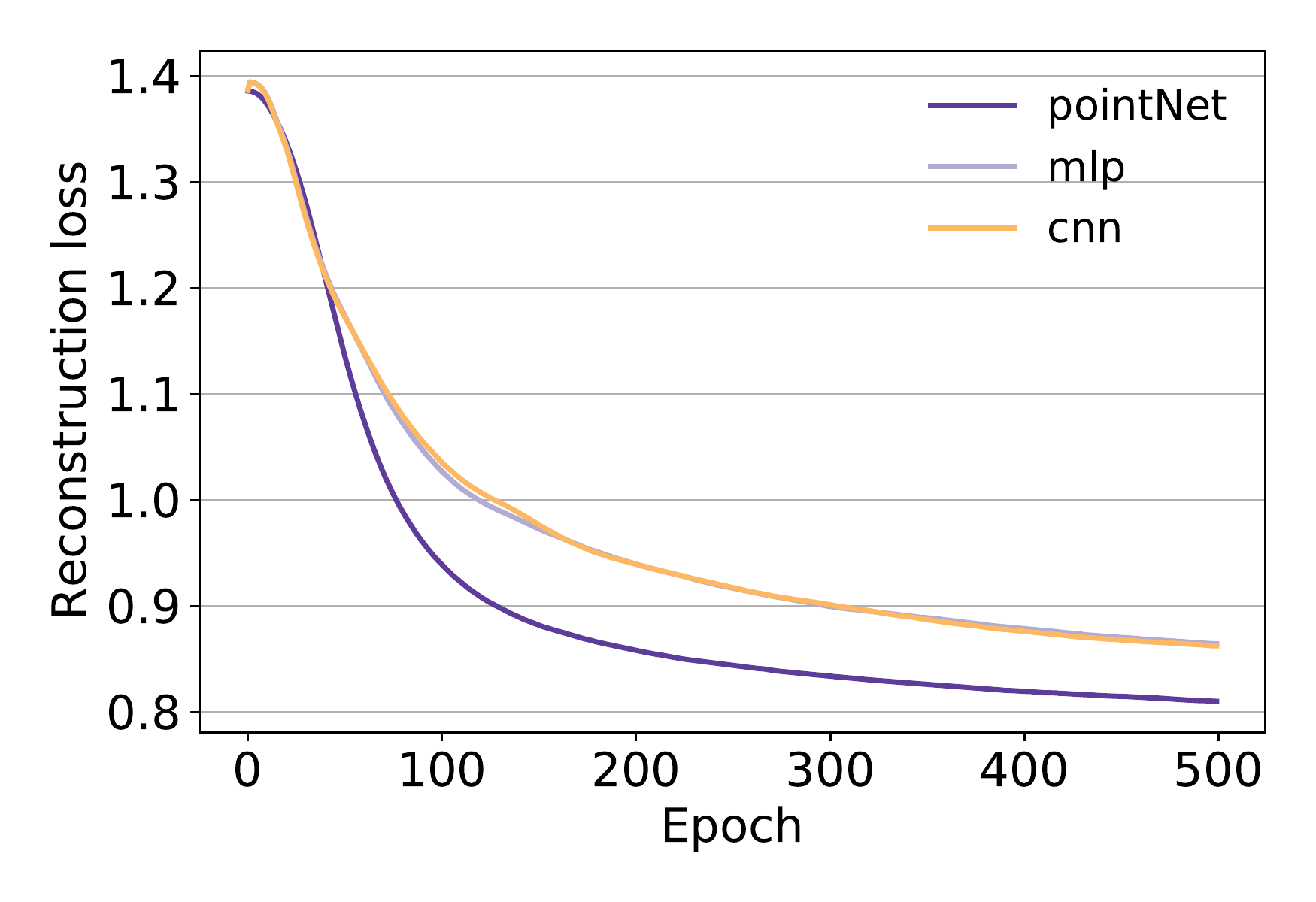}
		\caption{Reconstruction loss (Cora)}\label{fig:cora_r_loss}	
	\end{subfigure}
    \hfill
	\begin{subfigure}[t]{0.32\linewidth}
		\includegraphics[trim=0mm 7mm 0mm 0mm,clip,width=1.05\linewidth]{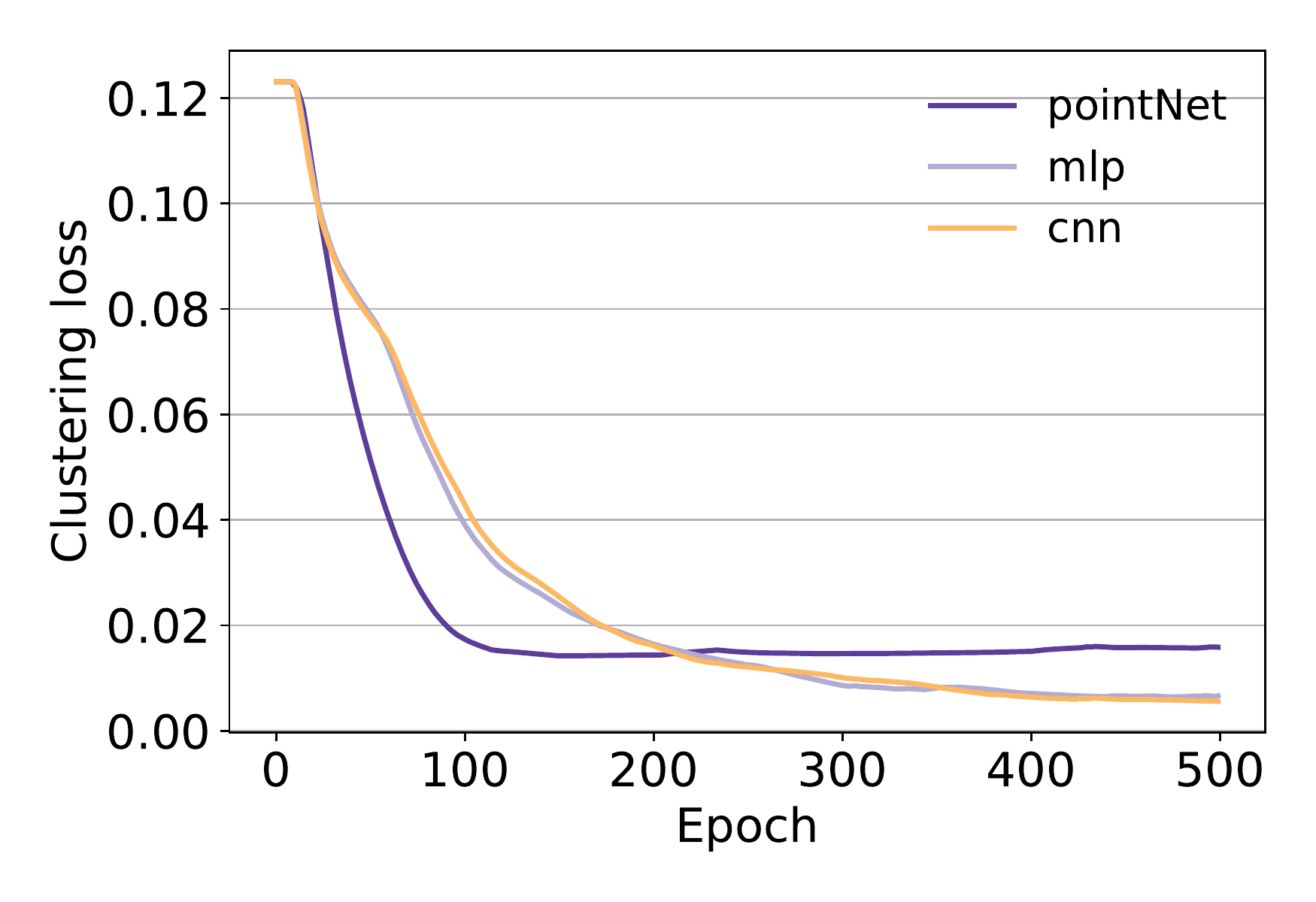}
		\caption{Clustering loss (Cora)}\label{fig:cora_c_loss}		
	\end{subfigure}
	\caption{Total loss, reconstruction loss and clustering loss of PointSpectrum with PointNetST, MLP and CNN-based networks in the encoder solving the clustering task on the Cora dataset (mean values over 10 runs). \textit{The set equivariant PointNetST helps the model converge faster than the MLP and CNN variants.}}	
	\label{fig:convergence}
\end{figure*}

\noindent\textbf{PointNetST.} In this work, PointNetST~\cite{segol2019universal} is employed as the permutation equivariant neural network architecture. While different choices can be made, such as a deep self-attention network~\cite{wang2018non} or the constructions of \cite{keriven2019universal} and~\cite{sannai2019universal}, PointNetST is preferred as it is provably a universal approximator over the space of equivariant functions~\cite{segol2019universal} and can be implemented as an arbitrarily deep neural network with the following form:
\begin{equation}
    f(\mathbf{X}) = \phi_m \circ \sigma \dots \circ \sigma \circ \phi_1(\mathbf{X})
\end{equation}
where $\sigma$ is a non-linearity such as ReLU and $\phi_i$, $i=1, ...,m$, is the DeepSet layer~\cite{zaheer2017deep}:
\begin{equation}
\label{eq:equivariant_transformation}
    \phi_i(\mathbf{X}) = \mathbf{X}\mathbf{W}_1 + \frac{1}{n}\mathbf{1}\mathbf{1}^T\mathbf{X}\mathbf{W}_2 + \mathbf{1}\mathbf{w}
\end{equation}
with $\mathbf{W}_1, \mathbf{W}_2 \in \mathbb{R}^{l_i \times l_{i+1}}$ and $\mathbf{w} \in \mathbb{R}^{l_{i+1}}$ being the layer's parameters. \textit{Remark:} while Eq.~\ref{eq:equivariant_transformation} is a generic form of an equivariant transformation, it is noted that PointNetST contains only a single layer with non-zero $\mathbf{W}_2$.

\subsection{PointSpectrum}
\label{sec:PointSpectrum}

This work proposes PointSpectrum, an encoder-decoder architecture for GRL that consists of the following main components: (i) Laplacian smoothing of the feature matrix, (ii) PointNetST as the permutation equivariant neural network of the encoder, and (iii) a clustering module  alongside the decoder. A schematic representation of PointSpectrum is depicted in Figure~\ref{fig:pointSpectrum_arch}.

\noindent \textbf{Input.} The input $\mathbf{\bar{X}}$ of the encoder-decoder network is the $k$-order graph convolution of node feature matrix $\mathbf{X}$, which is computed using a Laplacian filter $\mathbf{H}$ as described in Section~\ref{sec:smoothing}. The larger the convolution order $k$, the deeper node-wise interactions the model can capture; $k$ is a hyper-parameter of the model. 

\noindent \textbf{Encoder.} The encoder is the permutation equivariant PointNetST, which generates the node embeddings $\mathbf{Z}$, as described in Section~\ref{sec:equivariance}. The embeddings are fed to two individual modules: the decoder and the ClusterNet.

\noindent\textbf{Decoder:} The aim of the decoder is to reconstruct a pairwise similarity value between the computed node embeddings based on $\mathbf{\bar{X}}$. Different choices for the reconstruction loss function can be made (e.g. minimum squared error~\cite{wang2019attributed} or noise-contrastive binary cross entropy~\cite{velivckovic2018deep}). However, as connectivity is incorporated in the smoothed signal, we use a pairwise decoder and cross entropy with negative sampling as the loss function:
\begin{equation}\label{eq:reconstruction-loss}
    L_r = - \sum_{i \in \mathcal{V}}\left[\sum_{(i,j) \in \mathcal{E}}\log(\mathbf{z}_i\mathbf{z}_j) + \sum_{k \in \mathcal{N}_i}\log(1 - \mathbf{z}_i\mathbf{z}_k) \right]
\end{equation}
where $\mathcal{N}_i$ are the negative samples (i.e., non-existing edges) for node $i$.

\noindent \textbf{ClusterNet} is a differentiable clustering module, which learns to assign nodes to clusters to better separate them in the embedding space~\cite{wilder2019end}. ClusterNet learns a distribution of soft assignments $\mathbf{Q}$, where $q_{ij}$ expresses the probability of node $i$ belonging to cluster $j$, by optimizing KL-divergence loss function:
\begin{equation}\label{eq:clustering-loss}
    L_c = KL(P||Q) = \sum_i \sum_j p_{ij} \cdot \log\frac{p_{ij}}{q_{ij}}
\end{equation}
where $\mathbf{P}$ is a target distribution (updated in every epoch or using different intervals) that emphasizes the more ``confident'' assignments~\cite{wang2019attributed}:
\begin{equation}
    p_{ij} = \frac{q_{ij}^2/\sum_i q_{ij}}{\sum_j(q_{ij}^2/\sum_i q_{ij})}
\end{equation}

Having computed $\mathbf{Q}$, cluster centers can be extracted directly by averaging the soft assignments for each cluster.
Overall, PointSpectrum optimizes the following joint loss of the Decoder and ClusterNet
\begin{equation}\label{eq:joint-loss}
    L = \alpha\cdot  L_r + \beta\cdot  L_c
\end{equation}
with $\alpha,\beta$ being hyper parameters that control the importance of each component.

\section{Experimental Results}
\label{sec:results}

In this section, we demonstrate the efficiency of PointSpectrum in benchmark GRL datasets and tasks. We first present the datasets, the experimental setup, and the baseline methods we compare against to (Section~\ref{sec:datasets&exp}). Then, we provide experimental evidence for the gains that are introduced by the equivariance component of PointSpectrum over the traditional deep learning architectures in terms of efficiency (Sections \ref{sec:korder}), complexity (Section~\ref{sec:convergence}) and robustness (Section~\ref{sec:permutations}). Finally, we compare the performance of PointSpectrum against baseline and state-of-the art methods in GRL (Section~\ref{sec:baseline_comparison}) and provide a qualitative visual analysis (Section~\ref{sec:visual}).

\subsection{Datasets \& experimental setup}
\label{sec:datasets&exp}

\noindent\textbf{Datasets.} We evaluated the performance of PointSpectrum on three widely used benchmark citation network datasets, namely Cora~\cite{mccallum2000automating}, Citeseer~\cite{giles1998citeseer} and Pubmed~\cite{namata2012query}. The statistics of these data sources are presented and further discussed in Appendix~\ref{sec:datasets}. 

\begin{table*}[t]
	\begin{adjustbox}{width=\textwidth}
	\begin{small}
	\begin{tabular}{c}
	     \\
    	\begin{tabular}{l}
    	\textbf{Method} \\
    	\hline \\
    	MLP \\
    	\\
    	\hline \\
    	CNN \\
    	\\
    	\hline \\
    	PointNetST \\
    	\\
    	\end{tabular}
    \end{tabular}
	~
	\begin{tabular}{c}
	    \textbf{Cora} \\
    	\begin{tabular}{|l l l l}
    		 ACC & NMI & ARI & F1\\
    		\hline \\
    		0.275 & 0.247 & 0.170 & 0.145\\
    		(0.583) & (0.379) & (0.363) & (0.484)\\
    		\hline \\
    		0.301 & 0.090 & 0.514 & 0.208 \\
    		(0.571) & (0.417) & (0.346) & (0.489) \\
    		\hline \\
    		\textbf{0.625} & \textbf{0.431} & \textbf{0.385} & \textbf{0.538} \\
    		(0.715) & (0.528) & (0.493) & (0.693) \\
    	\end{tabular}
    \end{tabular}
	~
	\begin{tabular}{c}
	    \textbf{Citeseer} \\
    	\begin{tabular}{|l l l l}
    		ACC & NMI & ARI & F1\\
    		\hline \\
    		0.308 & 0.053 &  0.043 & 0.272 \\
    		(0.445) & (0.199) & (0.172) & (0.400) \\
    		\hline \\
    		0.362 & 0.098 & 0.091 & 0.314 \\
    		(0.423) & (0.197) & (0.175) & (0.395) \\
    		\hline \\
    		\textbf{0.537} & \textbf{0.316} & \textbf{0.271} & \textbf{0.463} \\
    		(0.703) & (0.430) & (0.451) & (0.613) \\
    	\end{tabular}
    \end{tabular}
	~
	\begin{tabular}{c}
	    \textbf{Pubmed} \\
    	\begin{tabular}{|l l l l}
    		ACC & NMI & ARI & F1\\
    		\hline \\ 
    		0.433 & 0.022 & 0.002 & 0.269\\
    		(0.638) & (0.210) & (0.212) & (0.639)\\
    		\hline \\
    		0.457 & 0.063 & 0.018 & 0.323\\
    		(0.684) & (0.263) & (0.298) & (0.673)\\
    		\hline \\
    		\textbf{0.691} & \textbf{0.298} & \textbf{0.307} & \textbf{0.687}\\
    		(0.710) & (0.295) & (0.329) & (0.703) \\
    	\end{tabular}
	\end{tabular}
	\end{small}
	\end{adjustbox}
	\caption{Clustering results based on the true labels. The input is randomly permuted in every epoch during training. The reported metrics result from the best clustering assignments of the original (not permuted) input. The results in the parenthesis refer to the models' performance when trained on the original input. \textit{PointSpectrum can capture data permutations on which it has not been explicitly trained, whereas the MLP/CNN variants perform poorly.}}
	\label{table:permutation}
\end{table*}

\noindent\textbf{PointSpectrum setup.} For all evaluation tasks, a PointSpectrum model is used for the encoder with a single PointNetST layer of dimension $l=100$ (which is also the embedding dimension). For the clusterNet module, centers are randomly initialized and correspond to the number of distinct labels in each dataset. Last, weights are initialized according to~\cite{he2015delving} (He initialization). Details for the hyperparameter tuning are given in Appendix~\ref{sec:hyperparams}.

\noindent\textbf{Baseline methods.} We compare PointSpectrum to several baseline methods (see details in Section~\ref{sec:related}), which we distinguish in four categories: (i) \textit{feature-only} (K-Means), (ii) \textit{traditional GRL} (DeepWalk, DNGR, TADW),  (iii) \textit{GNN-based} (VGAE, ARVGA - and their simpler derivatives - and DGI, GIC), and (iv) \textit{Spectral} (DAEGC, AGC, AGE). In Section~\ref{sec:baseline_comparison}, we compare against all these methods on clustering, and only the most prevalent ones on link prediction. Results are obtained directly from~\cite{mavromatis2020graph}; in particular, for AGE, K-means is used as the clustering method (instead of spectral clustering as in~\cite{cui2020adaptive}) to enable a fair comparison. In Sections~\ref{sec:korder}--\ref{sec:permutations}, the reported results correspond to the node clustering task, since it is the most natural task for PointSpectrum, considering its architecture.

\subsection{Efficiency of set equivariance}
\label{sec:korder}

We investigate the efficiency of using a set equivariant network (PointNetST) along with Laplacian smoothing ($\mathbf{\bar{X}}$), by comparing the PointSpectrum architecture (Figure~\ref{fig:pointSpectrum_arch}) against two variants with MLP and CNN neural networks in the encoder. Figure~\ref{fig:different_k_s} shows the results for the different encoder types and different convolution orders ($k$) for the clustering task on the Cora dataset. \footnote{
The corresponding results on the Citeseer and Pubmed datasets can be found in Appendix~\ref{sec:conv_order_appendix}; on Citeseer PointNetST performs even better than on Cora, while on Pubmed it performs similarly to MLP/CNN variants (due to the small number of features and simpler graph structure).}

\textit{PointNetST achieves higher performance, is more robust (lower variance), and has less fluctuations with respect to the convolution order than the MLP and CNN variants.}

This prevalence of PointNetST suggests that set equivariant networks are able to capture richer information when combined with Laplacian smoothing, and could be good candidates for replacing the conventional encoders of other Spectral methods as well (e.g., DAEGC, AGC, AGE).

\subsection{Set equivariance vs. training convergence}
\label{sec:convergence}

Set equivariant networks can offer multi-faceted benefits. As shown in~\cite{zaheer2017deep, segol2019universal} and in the results of Section~\ref{sec:korder}, they can capture richer structural information compared to the traditional MLP/CNN variants. In addition to these benefits, here we show that they also aid in training efficiency and computation complexity. Figure~\ref{fig:convergence} depicts the value of the loss function during the model's training for the original PointNetST and the MLP/CNN variants. \footnote{Similar findings hold for the Citeseer and Pubmed datasets; see Appendix~\ref{sec:convergence_appendix}}

\textit{In both components of the loss function (reconstruction, clustering) and the joint loss, PointNetST helps PointSpectrum to converge faster than the MLP/CNN variants}.

Remark: PointNetST achieves a lower loss value in overall (Figure \ref{fig:cora_loss}). While the clustering loss is slightly lower for the MLP/CNN variants (Figure~\ref{fig:cora_c_loss}) this difference is infinitesimal (second decimal point) compared to the reconstruction loss term (Figure~\ref{fig:cora_r_loss}). 

\begin{table*}[t]
	\begin{adjustbox}{width=\textwidth}
	\begin{small}
	\begin{tabular}{c}
	     \\
    	\begin{tabular}{l}
    	\textbf{Method} \\
    	\hline \\
    	K-Means \\
    	\hline \\
    	DeepWalk \\
    	DNGR \\
    	TADW \\
    	\hline \\
    	GAE/VGAE \\
    	ARGA/ARVGA \\
    	DGI \\
    	GIC \\
    	\hline \\
    	DAEGC \\
    	AGC \\
    	AGE \\
    	PointSpectrum (ours) \\
    	\end{tabular}
    \end{tabular}
	~
	\begin{tabular}{c}
	    \textbf{Cora} \\
    	\begin{tabular}{|l l l l}
    		 ACC & NMI & ARI & F1 \\
    		\hline \\
    		0.492 & 0.321 & 0.230 & 0.368\\
    		\hline \\
    		0.484 & 0.327 & 0.243 & 0.392 \\
    		0.419 & 0.318 & 0.142 & 0.340 \\
    		0.560 & 0.441 & 0.332 & 0.481 \\
    		\hline \\
    		0.609 & 0.436 & 0.347 & 0.609 \\
    		0.711 & 0.526 & 0.495 & 0.693 \\
    		0.713 & \textbf{0.564} & 0.511 & 0.682 \\
    		0.725 & 0.537 & 0.508 & 0.694 \\
    		\hline \\
    		0.704 & 0.528 & 0.496 & 0.682 \\
    		0.689 & 0.537 & 0.487 & 0.656 \\
    		0.712 & \underline{0.559} & - & 0.682 \\
    		\textbf{\underline{0.736}} & 0.529 & \textbf{\underline{0.516}} & \textbf{\underline{0.711}} \\
    	\end{tabular}
    \end{tabular}
	~
	\begin{tabular}{c}
	    \textbf{Citeseer} \\
    	\begin{tabular}{|l l l l}
    		ACC & NMI & ARI & F1 \\
    		\hline \\
    		0.540 & 0.305 & 0.279 & 0.409 \\
    		\hline \\
    		0.337 & 0.088 & 0.092 & 0.270 \\
    		0.326 & 0.180 & 0.044 & 0.300 \\
    		0.455 & 0.291 & 0.228 & 0.414 \\
    		\hline \\
    		0.408 & 0.176 & 0.124 & 0.372 \\
    		0.581 & 0.338 & 0.301 & 0.525 \\
    		0.688 & 0.444 & 0.450 & \textbf{0.657} \\
    		0.696 & \textbf{0.453} & \textbf{0.465} & 0.654 \\
    		\hline \\
    		0.672 & 0.397 & 0.410 & \underline{0.636} \\
    		0.670 & 0.411 & 0.419 & 0.625 \\
    		0.569 & 0.348 & - & 0.544 \\
    		\textbf{\underline{0.703}} & \underline{0.430} & \underline{0.451} & 0.613 \\
    	\end{tabular}
    \end{tabular}
	~
	\begin{tabular}{c}
	    \textbf{Pubmed} \\
    	\begin{tabular}{|l l l l}
    		ACC & NMI & ARI & F1 \\
    		\hline \\
    		0.398 & 0.001 & 0.002 & 0.195 \\
    		\hline \\
    		0.684 & 0.279 & 0.299 & 0.670 \\
    		0.458 & 0.155 & 0.054 & 0.467 \\
    		0.511 & 0.001 & 0.001 & 0.335 \\
    		\hline \\
    		0.672 & 0.277 & 0.279 & 0.660 \\
    		0.690 & 0.305 & 0.306 & 0.678 \\
    		0.533 & 0.181 & 0.166 & 0.186 \\
    		0.673 & 0.319 & 0.291 & 0.704 \\
    		\hline \\
    		0.671 & 0.266 & 0.278 & 0.659 \\
    		0.698 & 0.316 & 0.319 & 0.404 \\
    		- & - & - & - \\
    		\textbf{\underline{0.776}} & \textbf{\underline{0.375}} & \textbf{\underline{0.444}} & \textbf{\underline{0.768}} \\
    	\end{tabular}
	\end{tabular}
	\end{small}
	\end{adjustbox}
	\caption{Clustering results based on the true labels. Horizontal lines discriminate feature-only, traditional GRL, GNN-based and Spectral methods. \underline{Underlined} values indicate the best results among the spectral methods, and \textbf{bold} values the best results among all methods.}
	\label{table:clustering}
\end{table*}

\subsection{Performance on permuted data}
\label{sec:permutations}

As already shown, set equivariance offers performance and computational efficiency. In this section we focus on the main characteristic of set equivariance: its robustness on permuted inputs (which can be considered as a specific type of noisy/corrupted input data). To demonstrate this, we train PointSpectrum by randomly permuting the rows of $\mathbf{\bar{X}}$ (input data) in every epoch. Then, the trained model is evaluated on the original $\mathbf{\bar{X}}$.

Table~\ref{table:permutation} presents the results of this experiment, as well as the initial results without permuted data during training (in parentheses). PointNetST achieves the highest performance, but more importantly, the drop in performance due to the corrupted input is significantly smaller compared to the drop in the MLP/CNN variants. This highlights that \textit{PointSpectrum can capture data permutations on which it has not been explicitly trained} (e.g., in case of graph isomorphism). On the contrary, the MLP/CNN variants perform poorly on permuted data (see, e.g., the NMI/ARI metrics in the Citeseer and Pubmed datasets).

\subsection{Comparison against baselines}
\label{sec:baseline_comparison}

In this section, we compare PointSpectrum's efficiency in clustering and link prediction tasks against baseline and state-of-the-art methods. 

We would like to stress that the main goal of this work is to introduce set equivariance in Laplacian smoothing (Spectral) GRL methods, and demonstrate the benefits it can bring. Hence, we do not extensively emphasize on the hyperparameter optimization of PointSpectrum. Here, we compare its performance against baselines for completeness and for demonstrating its efficiency compared to the state-of-the-art in GRL. Nevertheless, the tested PointSpectrum implementation still outperforms Spectral methods, and achieves top or near top performance in the evaluation tasks.

\noindent\textbf{Clustering}: 
The goal in clustering is to group similar nodes into $m$ classes based on the computed embeddings. Similar to related literature, the number of classes is given, and in the evaluation the labels provided by the datasets are used. In Table~\ref{table:clustering} we report the best PointSpectrum results out of 10 experiment runs for 4 metrics: Accuracy (ACC), Normalized Mutual Information (NMI), Adjusted Randomized Index (ARI) and Macro-F1 score (F1).

Focusing first on the Spectral methods (bottom rows of Table~\ref{table:clustering}), we see that \textit{the overall PointSpectrum performance (i.e., for the majority of metrics and datasets) is superior to Spectral methods}. When compared to all GRL methods, \textit{PointSpectrum achieves state-of-the-art performance} on most metrics in the Cora and Pubmed datasets, as well as on the accuracy metric in the Citeseer dataset (in which the GNN-based models GIC and DGI perform best for the other metrics).

\noindent\textbf{Link prediction}: In link prediction, some graphs edges are hidden to the model during training and its goal is to predict these hidden interactions based on the computed node embeddings. For this task $10\%$ of positive and negative edges are used as test and $20\%$ as the validation set. Table~\ref{table:lp} presents the model performance on link prediction (mean value and standard deviation over 10 runs) as measured by the Area Under the Curve (AUC) and Average Precision (AP) metrics.

\textit{PointSpectrum outperforms all baselines by a significant margin in Cora ($\sim 5.5\%$ ) and Pubmed ($\sim 1.5\%$) datasets, while in Citeseer it is the second best method after GIC (with less than $1\%$ margin; also note that PointSpectrum has much lower variance than GIC).}

\begin{table*}[hbtp]
	\begin{adjustbox}{width=\textwidth}
	\begin{small}
	\begin{tabular}{c}
	     \\
    	\begin{tabular}{l}
    	\textbf{Method} \\
    	\hline \\
    	DeepWalk \\
    	GAE/VGAE \\
    	ARGA/ARVGA \\
    	DGI \\
    	GIC \\
    	AGE \\
    	PointSpectrum (ours) \\
    	\end{tabular}
    \end{tabular}
	~
	\begin{tabular}{c}
	    \textbf{Cora} \\
    	\begin{tabular}{|l l}
    		 AUC & AP \\
    		\hline \\
    		$0.846 \pm 0.001$ & $0.885 \pm 0.000$ \\
    		$0.914 \pm 0.001$ & $0.926 \pm 0.001$ \\
    		$0.924 \pm 0.000$ & $0.932 \pm 0.000$ \\
    		$0.898 \pm 0.08$ & $0.897 \pm 0.1$ \\
    		\underline{$0.935 \pm 0.06$} & \underline{$0.933 \pm 0.07$} \\
    		$0.924 \pm 0.000$ & $0.932 \pm 0.000$ \\
    		$\mathbf{0.989 \pm 0.000}$ & $\mathbf{0.989 \pm 0.001}$ \\
    	\end{tabular}
    \end{tabular}
    ~
	\begin{tabular}{c}
	    \textbf{Citeseer} \\
    	\begin{tabular}{|l l}
    		AUC & AP \\
    		\hline \\
    		$0.805 \pm 0.002$ & $0.850 \pm 0.001$ \\
    		$0.908 \pm 0.002$ & $0.920 \pm 0.002$ \\
    		$0.924 \pm 0.000$ & $0.930 \pm 0.000$ \\
    		$0.955 \pm 0.1$ & $0.957 \pm 0.1$ \\
    		$\mathbf{0.970 \pm 0.05}$ & $\mathbf{0.968 \pm 0.05}$ \\
    		$0.924 \pm 0.000$ & $0.930 \pm 0.000$ \\
    		\underline{$0.966 \pm 0.000$} & \underline{$0.961 \pm 0.001$} \\
    	\end{tabular}
    \end{tabular}
	~
	\begin{tabular}{c}
	    \textbf{Pubmed} \\
    	\begin{tabular}{|l l}
    		AUC & AP\\
    		\hline \\
    		$0.842 \pm 0.000$ & $0.878 \pm 0.000$ \\
    		$0.964 \pm 0.000$ & $0.965 \pm 0.000$ \\
    		\underline{$0.968 \pm 0.000$} & \underline{$0.971 \pm 0.000$} \\
    		$0.912 \pm 0.06$ & $0.922 \pm 0.05$ \\
    		$0.937 \pm 0.03$ & $0.935 \pm 0.03$ \\
    		\underline{$0.968 \pm 0.000$} & \underline{$0.971 \pm 0.000$} \\
    		$\mathbf{0.987 \pm 0.001}$ &  $\mathbf{0.982 \pm 0.002}$ \\
    	\end{tabular}
	\end{tabular}
	\end{small}
	\end{adjustbox}
	\caption{Link prediction performance. Area Under the Curve (AUC) and Average Precision (AP) are reported. Best results on each dataset are shown in \textbf{bold}, and second best in \underline{underlined}.}
	\label{table:lp}
\end{table*}

\subsection{Qualitative Analysis}
\label{sec:visual}

We evaluate qualitatively PointSpectrum by visualizing the computed embeddings using t-SNE~\cite{van2008visualizing} for the Cora dataset in Figure~\ref{fig:cora_tsne} (see Appendix~\ref{sec:tsne_appendix} for visualizations on the Citeseer and Pubmed datasets).

\begin{figure}[htbp]
	\begin{subfigure}[t]{0.32\linewidth}
		\includegraphics[width=1\linewidth]{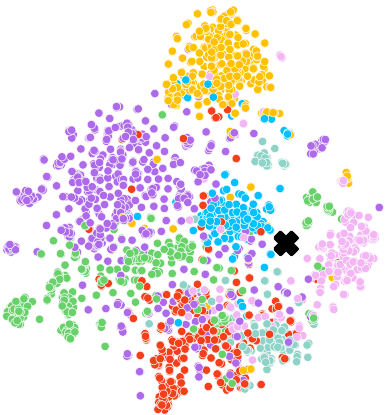}
		\caption{Initial}
	\end{subfigure}
    \hfill
	\begin{subfigure}[t]{0.32\linewidth}
		\includegraphics[width=1\linewidth]{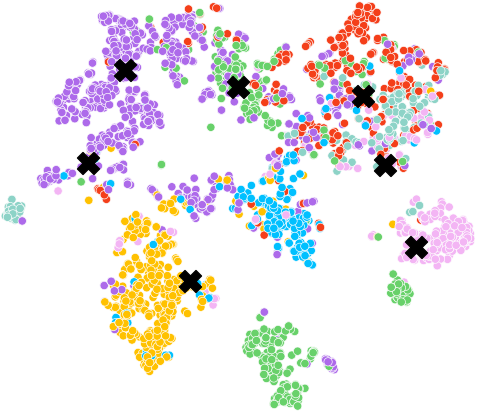}
		\caption{Intermediate}
	\end{subfigure}
    \hfill
	\begin{subfigure}[t]{0.32\linewidth}
		\includegraphics[width=1\linewidth]{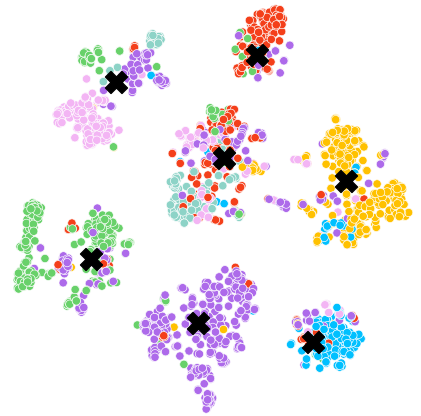}
		\caption{Final}
	\end{subfigure}
	\caption{PointSpectrum's training behavior on Cora dataset using t-SNE for visualization. ClusterNet's trainable centers are denoted as black `x' marks.}
	\label{fig:cora_tsne}
\end{figure}

On one hand, we observe that PointSpectrum separates well the nodes in the embedding space as the training proceeds. On the other hand, the ClusterNet component enables the PointSpectrum to learn the cluster centers as well (denotes as `x' marks), as it pushes them from a random initial point towards each group's center. Last, since training is an iterative process, node embeddings and cluster centers attract each other in turns, explaining the formulation of these distinct clusters.

\section{Related Work}
\label{sec:related}

GRL has gained a lot of attention due to the need for automated processes that can analyze large volumes of structured data, with graphs being the hallmark of such structures.

\noindent\textbf{Conventional graph embeddings}: The first efforts exploited well-known graph mechanisms to calculate node representations. DeepWalk~\cite{perozzi2014deepwalk} and Node2Vec~\cite{grover2016node2vec} utilize random walks to sample the graph and train a Word2Vec model on these samples to extract the embeddings. Also, TADW~\cite{yang2015network} applies non-negative matrix factorization on both the graph and node features to get a consistent partition. Last, DNGR~\cite{cao2016deep} employs denoising auto encoders to find low dimensional representations and then reconstruct the graph adjacency.

\noindent\textbf{GNN-based methods}: Graph Neural Networks are designed to capture graph structures, and thus have been used for learning node representations. VGAE~\cite{kipf2016variational} uses GCNs to form a variational autoencoder which learns to generate node embeddings, while ARVGA~\cite{pan2018adversarially} uses adversarial learning to train the graph auto encoder. DGI~\cite{velivckovic2018deep} leverages both local and global graph information for representation learning using contrastive learning, while GIC~\cite{mavromatis2020graph} extends DGI by forming node clusters to better separate nodes in the embedding space. In particular ClusterNet~\cite{wilder2019end} --the clustering process of GIC which is crucial for its superior performance-- is also incorporated in PointSpectrum aiding in performance and validating GIC's design. Although efficient, to capture deep graph interactions GNN methods are inevitably led to over smoothing, where node representations converge to indistinguishable vectors~\cite{chen2020measuring, zhou2020towards}.

\noindent\textbf{Spectral methods}: On the other hand, spectral methods exploit graph filters to perform high-order graph convolution at once, thus bypassing GNNs' over smoothing. AGC~\cite{zhang2019attributed} uses Laplacian filtering with spectral clustering to cluster nodes into groups, while AGE~\cite{cui2020adaptive} employs an auto encoder to produce node embeddings through Laplacian smoothing. Also, DAEGC~\cite{wang2019attributed} leverages an attentional network alongside soft-labeling for self supervision to construct the embeddings. 
While spectral methods address over smoothing, they have only used conventional neural networks (MLPs, CNNs) that cannot capture graph properties (e.g., equivariance) by design; they can only learn the structural information contained in the smoothed input signal. 

PointSpectrum --although a spectral method-- lies on the intersection of GNN-based and spectral methods, alleviating over smoothing through graph filtering and capturing structural information through set equivariant networks.

\section{Conclusion}
\label{sec:discussion}
PointSpectrum is the first work to introduce the set equivariance property (typically, a property of GNN-based methods) into spectral methods. Set equivariance is important when learning on graph data, since it is inherently designed to exploit the nature of unordered data. Our work was motivated by this, and our experimental results clearly demonstrated the performance benefits of using a set equivariant network (PointNetST) over the MLP or CNN layers that are used in spectral methods.

We deem PointSpectrum as an initial effort (or as a proof of concept) in the direction of integrating set equivariance with Laplacian smoothing. This is why we adopted a simple design for the model architecture, without exhaustively over-engineering its modules or tuning its hyperparameters. Nevertheless, and despite this simplicity, we have shown that PointSpectrum can achieve state-of-the-art results in benchmark datasets. This brings a positive message for the efficiency, applicability and generalizability of our approach to other spectral or more generic GRL methods.

In particular, we identify the following as promising directions for future research: 

\noindent\textbf{Extensions}: Set equivariant networks (e.g., DeepSet or PointNetST) can easily be introduced to existing spectral methods (e.g., AGC, AGE or DAEGC) by replacing the MLP or CNN layers that they use. A more challenging direction is the extension of the proposed approach to \textit{generative models}, such as VAE or GAN architectures.

\noindent\textbf{Generalization}: As shown, PointSpectrum performs well even under data permutations. A deeper understanding (experimental/theoretical) of its capacity to generalize on noisy, corrupted or unseen data, could provide further insights on the mechanics of using set equivariant methods on graphs, as well as lead to the design of more efficient GRL methods. 

\noindent\textbf{Unification}: PointSpectrum has a modular design, where a set equivariant network receives as input the smoothed matrix $\mathbf{\bar{X}}$. Unifying these two operations in a single component (e.g., a new GNN layer), if possible, could simultaneously aim at a higher performance and bypass over smoothing.


\bibliography{main}

\include{appendix}

\end{document}

%% file: appendix.tex
\clearpage
\appendix
\section{Appendix}
\label{Appendix}

\subsection{Datasets}
\label{sec:datasets}

\begin{table}[htbp]
	\begin{center}
    \begin{tabular}{l|llll}
		\bf Dataset  &\bf Nodes &\bf Edges &\bf Features & \bf Classes
        \\ \hline \\
		Cora & 2708 & 5429 & 1433 & 7 \\
	    Citeseer & 3327 & 4732 & 3703 & 6 \\
	    Pubmed & 19717 & 44338 & 500 & 3 \\
	\end{tabular}
	\end{center}
	\caption{Dataset specifics}
	\label{table:datasets}
\end{table}

Table~\ref{table:datasets} presents the statistics of the datasets used throughout the experimental process. These data contain a different number of labels, which are used as the oracle information to calculate the reported metrics (Accuracy, NMI, ARI and F1). All of them present a sparse graph structure, while Pubmed contains less rich information in terms of node features in comparison to Cora and Citeseer.

\subsection{Hyperparameter setup}
\label{sec:hyperparams}

\begin{table}[htbp]
	\begin{center}
	\begin{adjustbox}{width=\linewidth}
    \begin{tabular}{llllllll}
		\bf Dataset  &\bf k &\bf Dropout &\bf LR & \bf Epochs &\bf Dim & \textbf{$\alpha$ (value)} & \textbf{$\beta$ (value)}
        \\ \hline \\
		Cora & 5 & 0.2 & 0.01 & 500 & 100 & const (1) & const (2)\\
	    Citeseer & 1 & 0.2 & 0.01 & 500 & 100 & expdec (1) & exp (5)\\
	    Pubmed & 7 & 0.2 & 0.01 & 500 & 100 & const (1) & const (2)\\
	\end{tabular}
	\end{adjustbox}
	\end{center}
	\caption{Hyper-parameter values for different datasets. $\alpha$ and $\beta$ types are: constant (const), exponential increase (exp) and exponential decrease (expdec). k refers to convolution order,  LR to learning rate and Dim to encoder's dimensions.}
	\label{table:hyperparams}
\end{table}

To conduct the experiments and validate PointSpectrum's performance, hyper parameter tuning is needed. The values corresponding to each dataset are presented in Table~\ref{table:hyperparams}. It should be noted that the best value for convolution order is smaller for PointSpectrum when compared to other methods that employ graph filtering (8, 6 and 8 for Cora, Citeseer and Pubmed respectively). This showcases the fact that set equivariant networks can capture structural information more easily and thus they do not need the whole information to be presented explicitly.

Regarding hyper parameters $\alpha$ and $\beta$, various configurations were investigated. Specifically, we point out the below behaviors:
\begin{itemize}
    \item constant: the hyper parameter has a constant value throughout training
    \item linear: the hyper parameter linearly increases (decreases) in every epoch to reach a maximum (minimum) value, which is also provided by the user
    \item exponential: the hyper parameter exponentially increases (decreases) in every epoch to reach a maximum (minimum) value, which is also provided by the user. Specifically, given the maximum value $s$ and the number of epochs $e$ the following function is used: $f(x) = \exp{(\frac{\log{1+s}}{e} x)} - 1$ for a given epoch x. For the decreasing values, we sort this function's results in decreasing order.
\end{itemize}

\subsection{Ablation study}

\begin{table*}[t]
	\begin{adjustbox}{width=\textwidth}
	\begin{small}
	\begin{tabular}{c}
	     \\
    	\begin{tabular}{l}
    	\textbf{Method} \\
    	\hline \\
    	Conventional AE (MLP) \\
    	Conventional AE (CNN) \\
    	\hline \\
    	~~ + PointNetST \\
    	\hline \\
    	~~~~ + ClusterNet \\
    	\end{tabular}
    \end{tabular}
	~
	\begin{tabular}{c}
	    \textbf{Cora} \\
    	\begin{tabular}{|l l l l}
    		 ACC & NMI & ARI & F1 \\
    		\hline \\
    		0.578 & 0.414 & 0.333 & 0.526 \\
    		0.610 & 0.423 & 0.362 & 0.560 \\
    		\hline \\
    		0.645 & 0.482 & 0.428 & 0.561 \\
    		\hline \\
    		\textbf{0.736} & \textbf{0.529} & \textbf{0.516} & \textbf{0.711}
    	\end{tabular}
    \end{tabular}
	~
	\begin{tabular}{c}
	    \textbf{Citeseer} \\
    	\begin{tabular}{|l l l l}
    		ACC & NMI & ARI & F1 \\
    		\hline \\
    		0.480 & 0.215 & 0.198 & 0.428 \\
    		0.465 & 0.212 & 0.189 & 0.415 \\
    		\hline \\
    		0.673 & 0.393 & 0.407 & 0.595 \\
    		\hline \\
    		\textbf{0.703} & \textbf{0.430} & \textbf{0.451} & \textbf{0.613} \\
    	\end{tabular}
    \end{tabular}
	~
	\begin{tabular}{c}
	    \textbf{Pubmed} \\
    	\begin{tabular}{|l l l l}
    		ACC & NMI & ARI & F1 \\
    		\hline \\
    		0.607 & 0.248 & 0.206 & 0.596 \\
    		0.594 & 0.231 & 0.189 & 0.584 \\
    		\hline \\
    		0.531 & 0.207 & 0.127 & 0.539 \\
    		\hline \\
    		\textbf{0.776} & \textbf{0.375} & \textbf{0.444} & \textbf{0.768} \\
    	\end{tabular}
	\end{tabular}
	\end{small}
	\end{adjustbox}
	\caption{Ablation study.}
	\label{table:ablation}
\end{table*}

To validate the efficacy of PointSpectrum's components an ablation study is conducted. First a conventional autoencoder is tested using either an MLP or a CNN encoder. Then PointNetST substitutes the conventional autoencoder and last ClusterNet is also employed alongside reconstruction objective. As it can be seen in Table~\ref{table:ablation}, in all three datasets the holistic PointSpectrum model achieves the best results. Furthermore, regarding Cora and Citeseer both set equivariance (through PointNetST) and the clustering objective (through ClusterNet) increase the model's performance. However, for Pubmed although clustering is beneficial, set equivariance does not seem to help. This may relate to the low number of features when compared to the graph size, meaning that the information captured from the conventional neural networks is sufficient to characterize the data, while node features (and thus their permutations) are not that important. 

\begin{figure}[htbp]
	\begin{subfigure}[t]{0.49\linewidth}
		\includegraphics[width=1\linewidth]{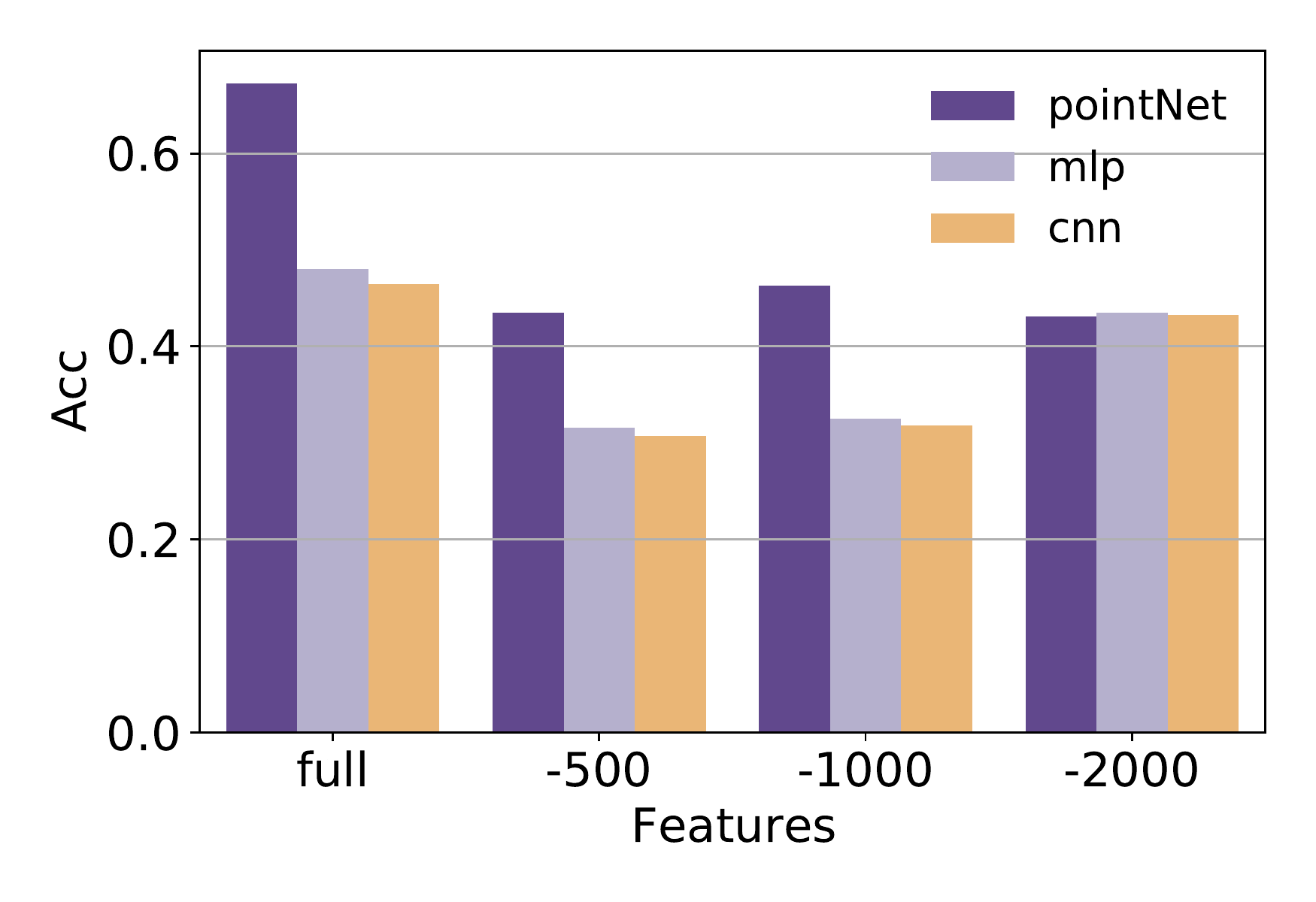}
		\caption{Accuracy (Cora)}\label{fig:features_acc}		
	\end{subfigure}
    \hfill
	\begin{subfigure}[t]{0.49\linewidth}
		\includegraphics[width=1\linewidth]{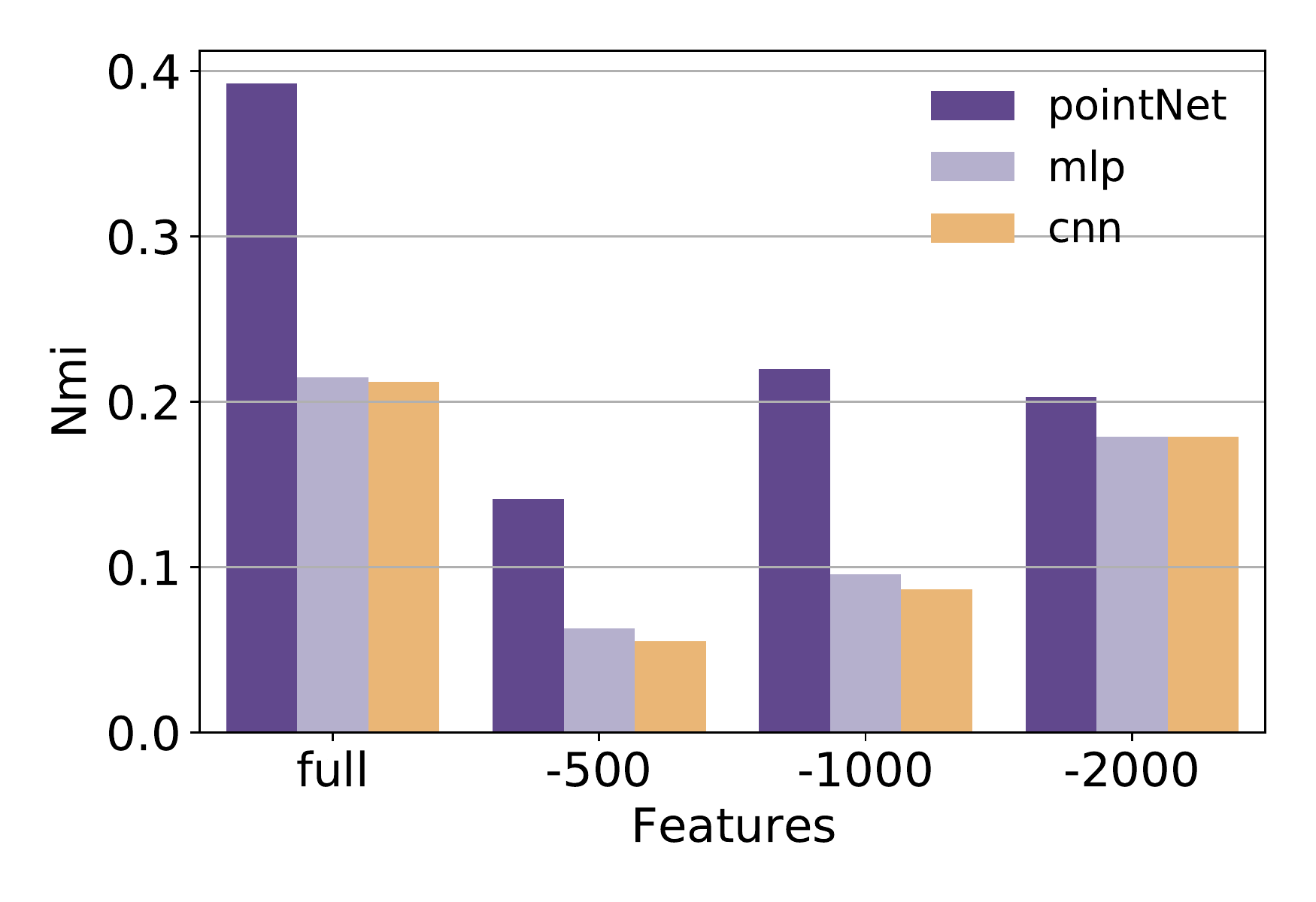}
		\caption{NMI (Cora)}\label{fig:features_nmi}	
	\end{subfigure}
    \hfill
	\\
	\begin{subfigure}[t]{0.49\linewidth}
		\includegraphics[width=1\linewidth]{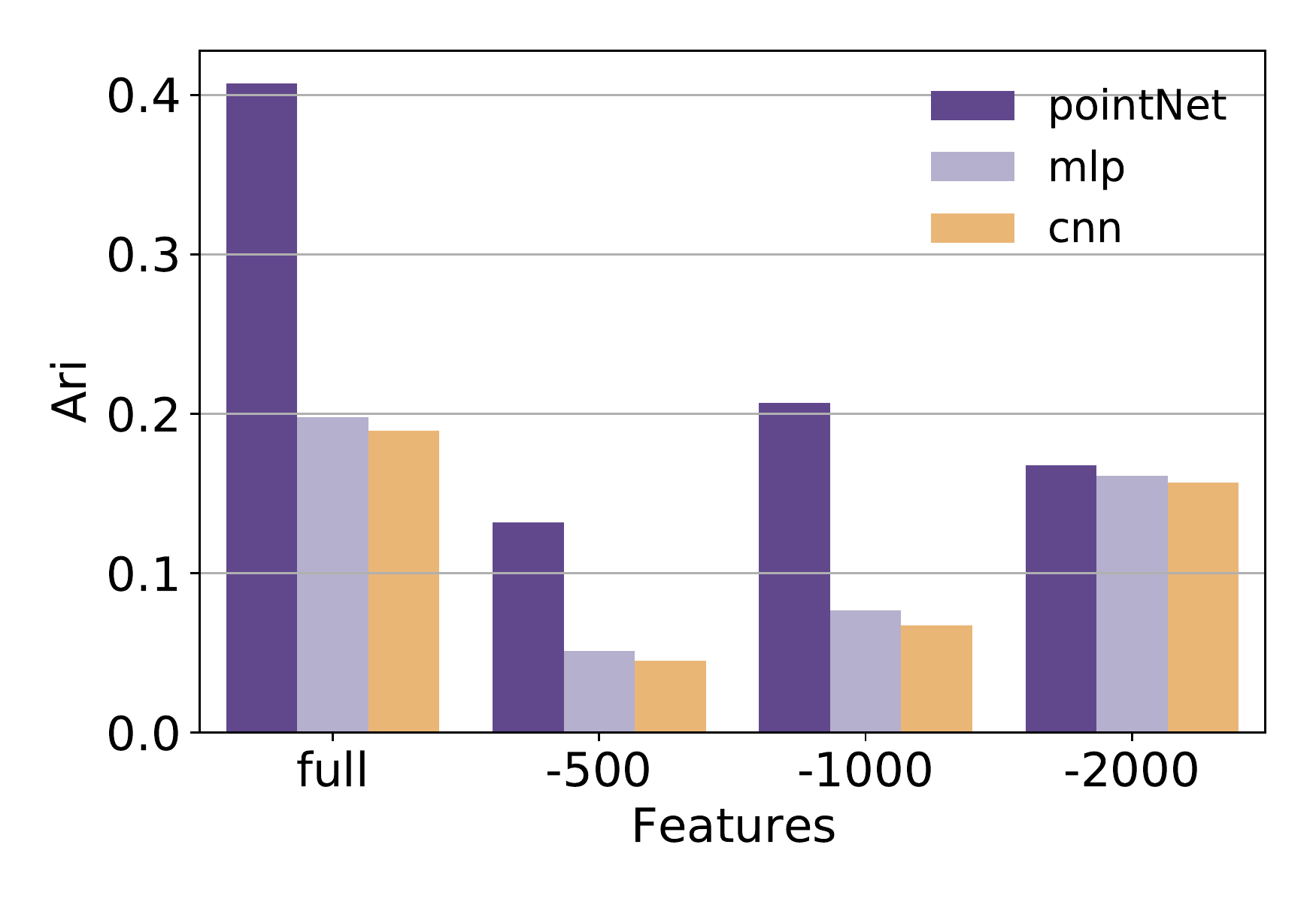}
		\caption{ARI (Citeseer)}\label{fig:features_ari}	
	\end{subfigure}
    \hfill
	\begin{subfigure}[t]{0.49\linewidth}
		\includegraphics[width=1\linewidth]{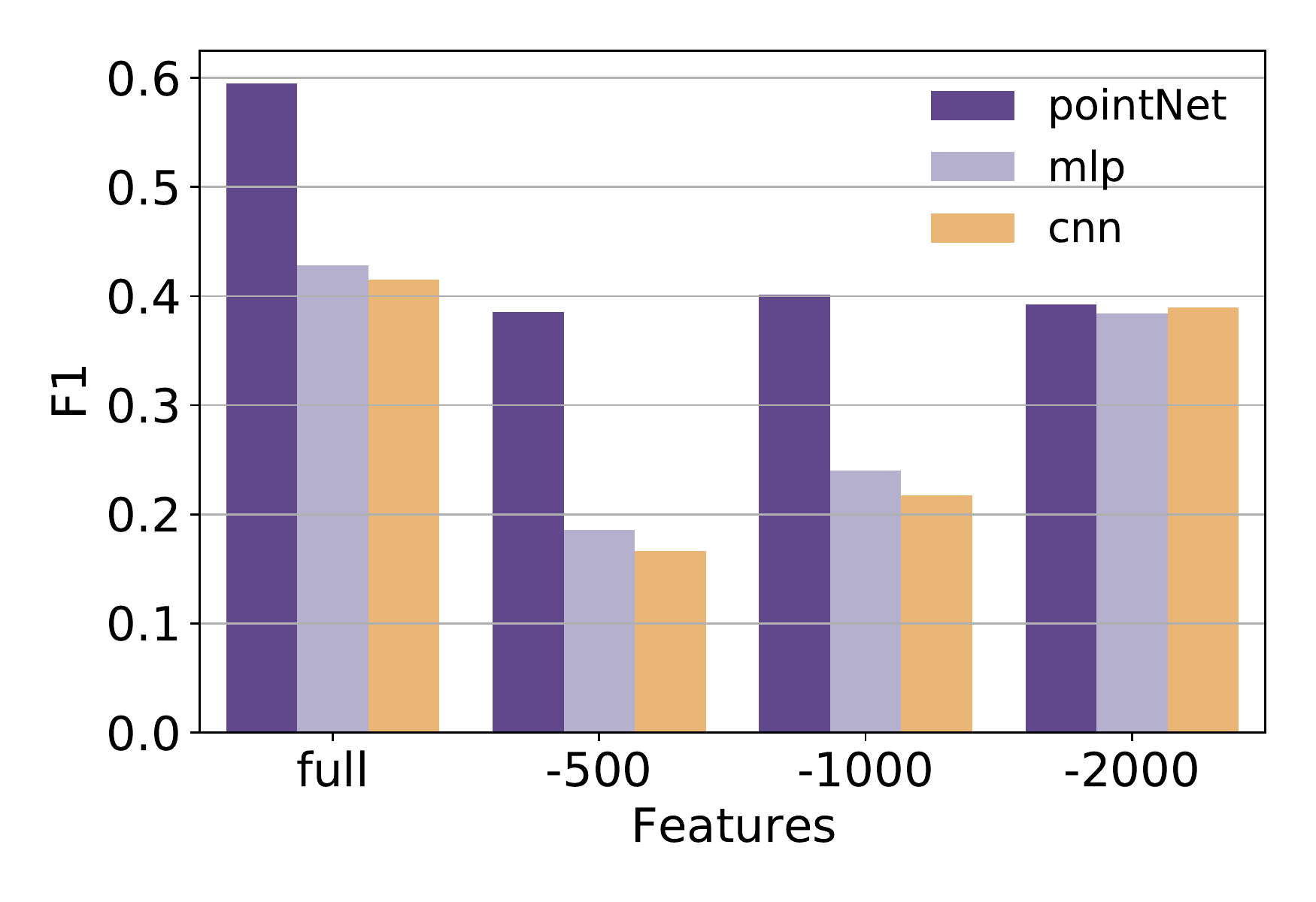}
		\caption{F1 (Citeseer)}\label{fig:features_f1}		
	\end{subfigure}
	\caption{PointNetST, MLP and CNN-based methods' performance trained and evaluated on reduced number of features (depicted as pointNet, mlp and cnn respectively.}
	\label{fig:feature_importance}
\end{figure}

To verify the above assumption we have also tested the three PointSpectrum variants on a reduced number of features. Specifically, Figure~\ref{fig:feature_importance} depicts this experiment on Citeseer dataset where the number of features is large enough to enable us different degrees of features reduction. Although performance deterioration is not analogous to features reduction, a general trend is shown: PointNetST is heavily dependent on features, while conventional neural networks seem to pull closer to their full-features performance no matter the reduction. However, this may come from conventional neural networks' tendency to overfit to the specific permutation, therefore a more thorough investigation is needed which is out of this work's scope.

\subsection{Additional visual results}
\label{sec:tsne_appendix}

Here the visual analyses of tSNE on Citeseer and Pubmed are presented. As depicted, PointSpectrum behaves similar to Cora, separating the embeddings as training proceeds and forcing cluster centers towards the centers of the node groups that it creates. More concretely, in the case of Citeseer, nodes are well separated into distinct clusters and therefore the trained ClusterNet centers match to the actual centers. For Pubmed though, node separation is not trivial and nodes appear to be similar to nodes of other clusters. We suppose that the reason behind this phenomenon is the small number of available features for PointSpectrum to exploit, when compared to Cora and Citeseer. 

\begin{figure}[htbp]
	\begin{subfigure}[t]{0.32\linewidth}
		\includegraphics[width=1\linewidth]{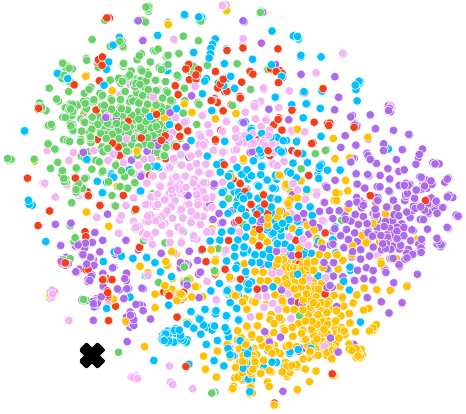}
	\end{subfigure}
    \hfill
	\begin{subfigure}[t]{0.32\linewidth}
		\includegraphics[width=1\linewidth]{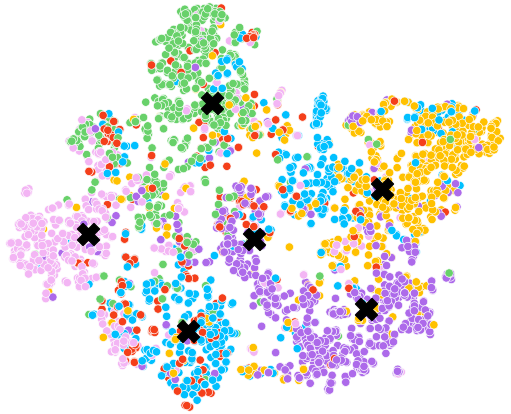}
	\end{subfigure}
    \hfill
	\begin{subfigure}[t]{0.32\linewidth}
		\includegraphics[width=1\linewidth]{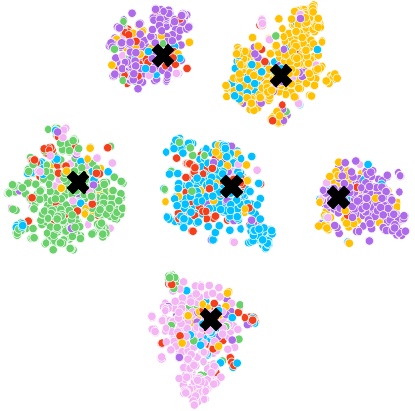}
	\end{subfigure}
	\\
	\begin{subfigure}[t]{0.32\linewidth}
		\includegraphics[width=1\linewidth]{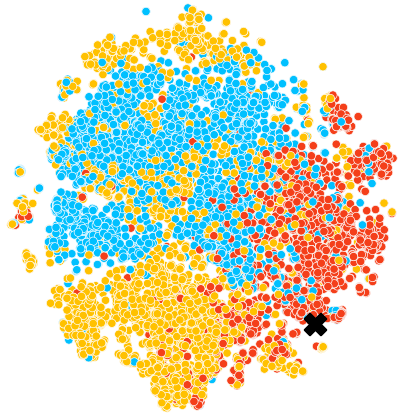}
		\caption{Initial}
	\end{subfigure}
    \hfill
	\begin{subfigure}[t]{0.32\linewidth}
		\includegraphics[width=1\linewidth]{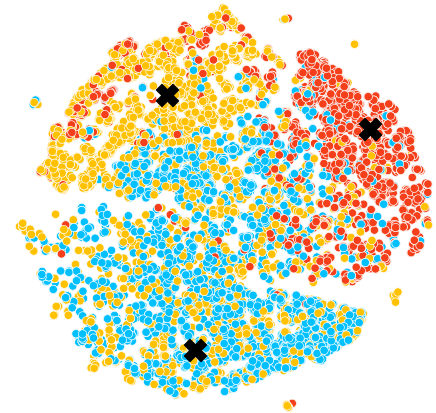}
		\caption{Intermediate}
	\end{subfigure}
    \hfill
	\begin{subfigure}[t]{0.32\linewidth}
		\includegraphics[width=1\linewidth]{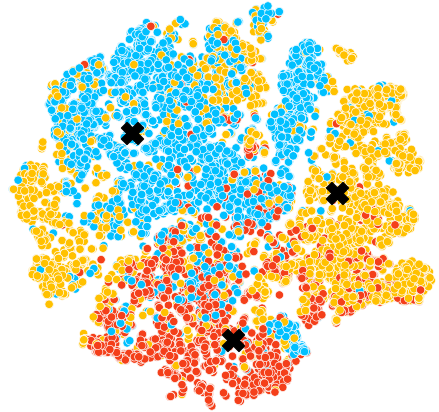}
		\caption{Final}
	\end{subfigure}
	\caption{PointSpectrum's training behavior on Citeseer (top) and Pubmed (bottom) dataset using t-SNE for visualization. ClusterNet's trainable centers are denoted as black `x' marks.}
	\label{fig:citeseer_tsne}
\end{figure}

\subsection{Influence of convolution order on Citeseer and Pubmed}
\label{sec:conv_order_appendix}

\begin{figure*}[hbtp]
	\begin{subfigure}[t]{0.32\linewidth}
		\includegraphics[width=1\linewidth]{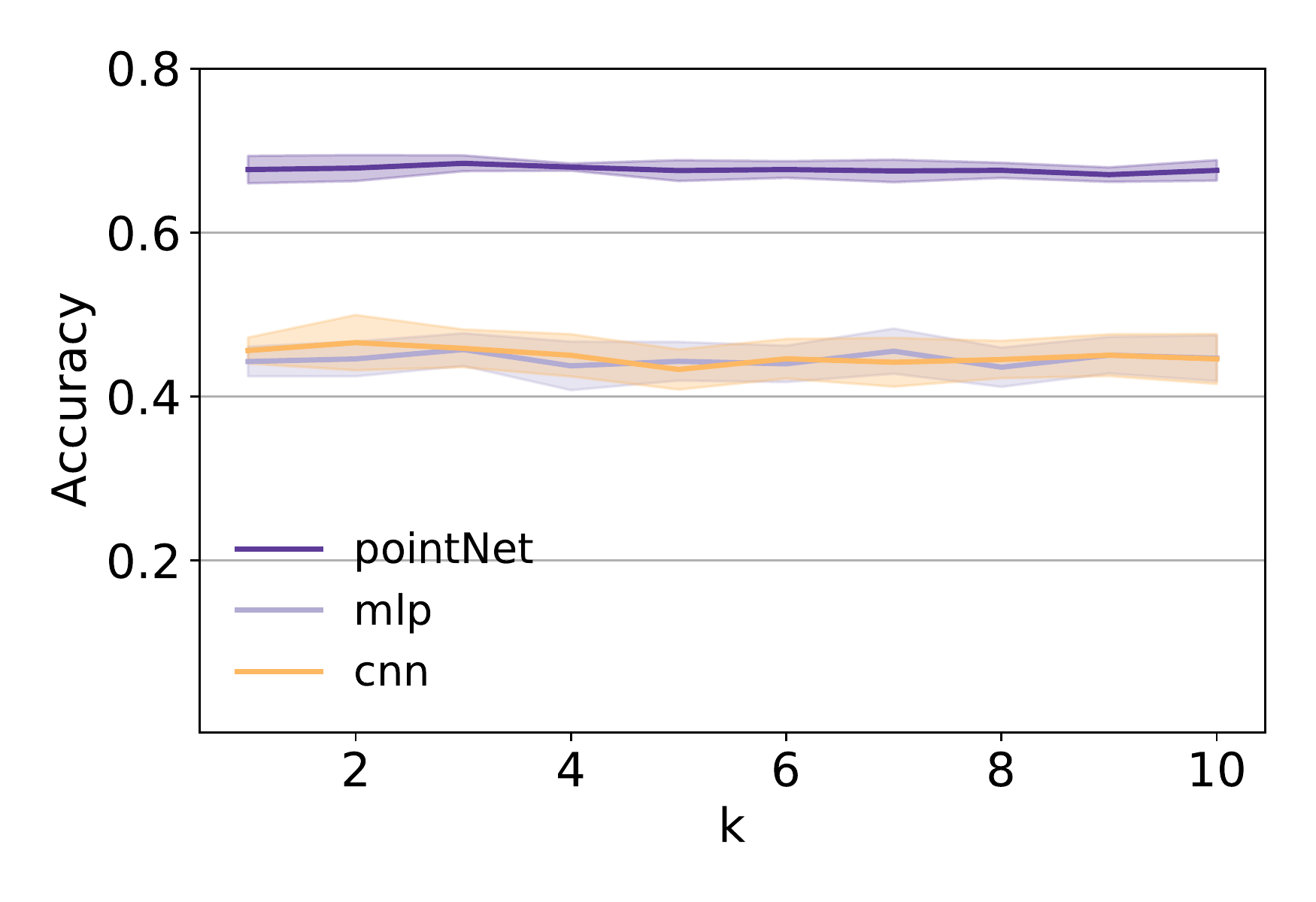}
    		\caption{Accuracy (Citeseer)}\label{fig:citeseer_acc}		
	\end{subfigure}
    \hfill
	\begin{subfigure}[t]{0.32\linewidth}
		\includegraphics[width=1\linewidth]{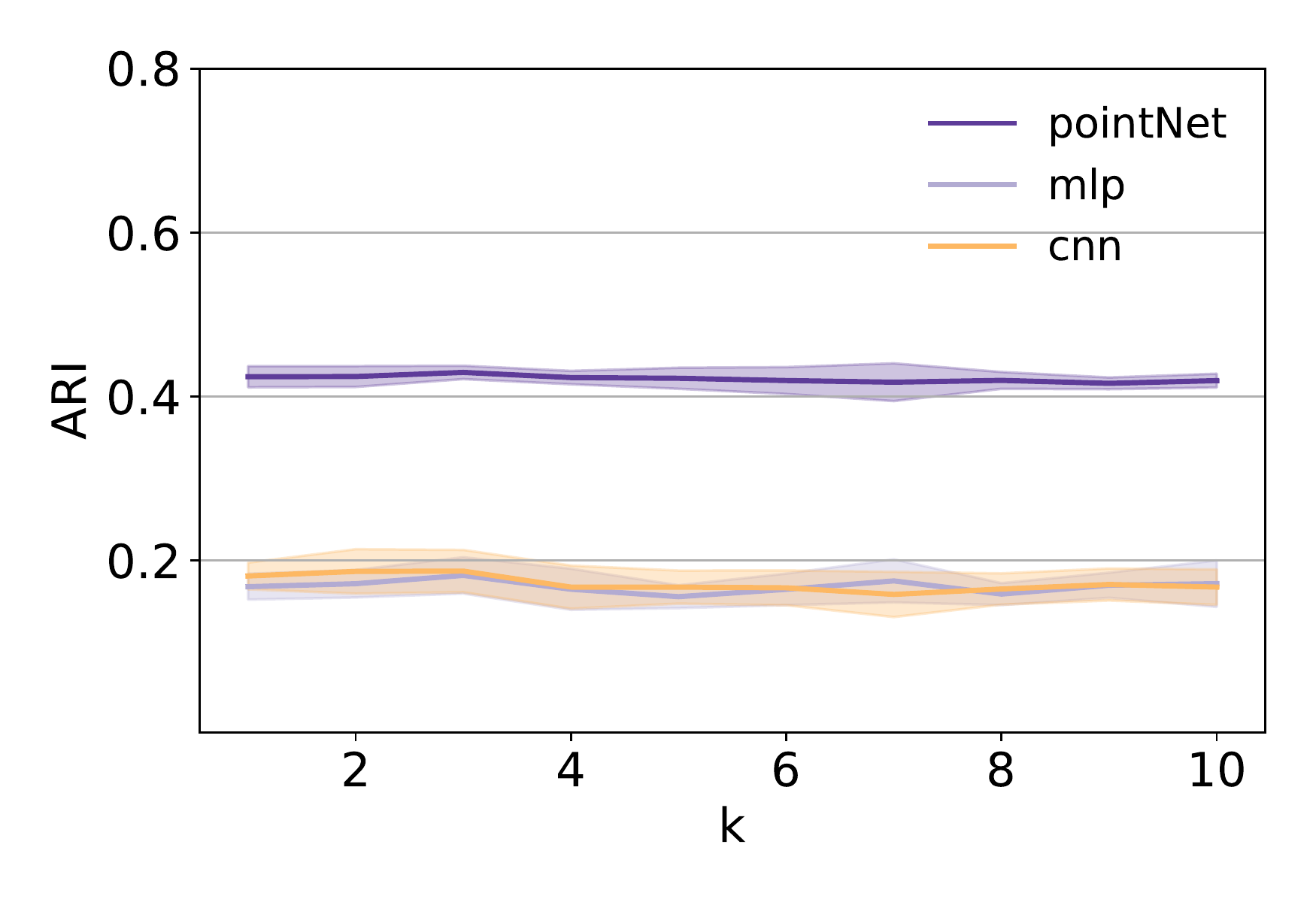}
		\caption{ARI (Citeseer)}\label{fig:citeseer_ari}	
	\end{subfigure}
    \hfill
	\begin{subfigure}[t]{0.32\linewidth}
		\includegraphics[width=1\linewidth]{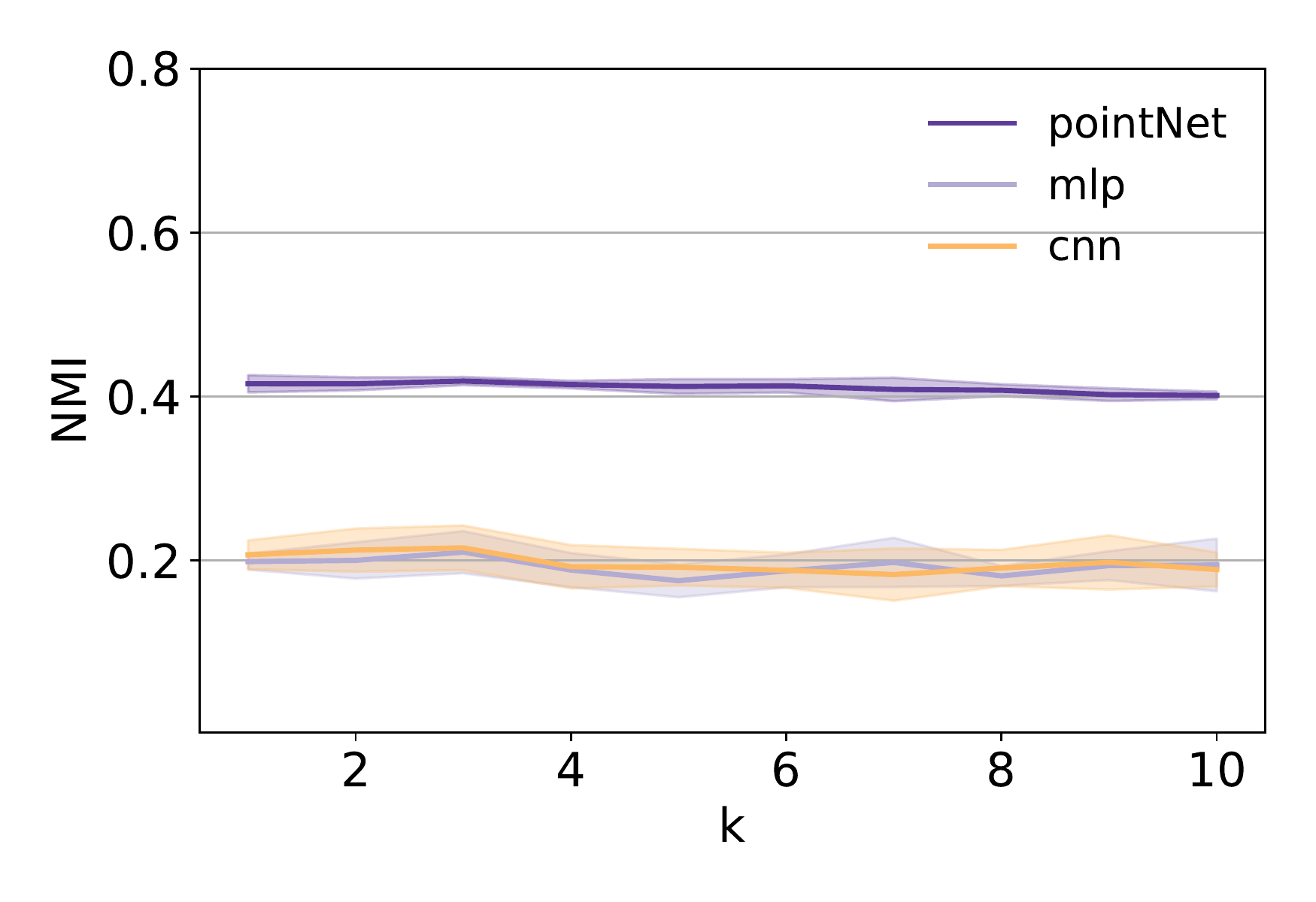}
		\caption{NMI (Citeseer)}\label{fig:citeseer_nmi}		
	\end{subfigure}
	\\
	\begin{subfigure}[t]{0.32\linewidth}
		\includegraphics[width=1\linewidth]{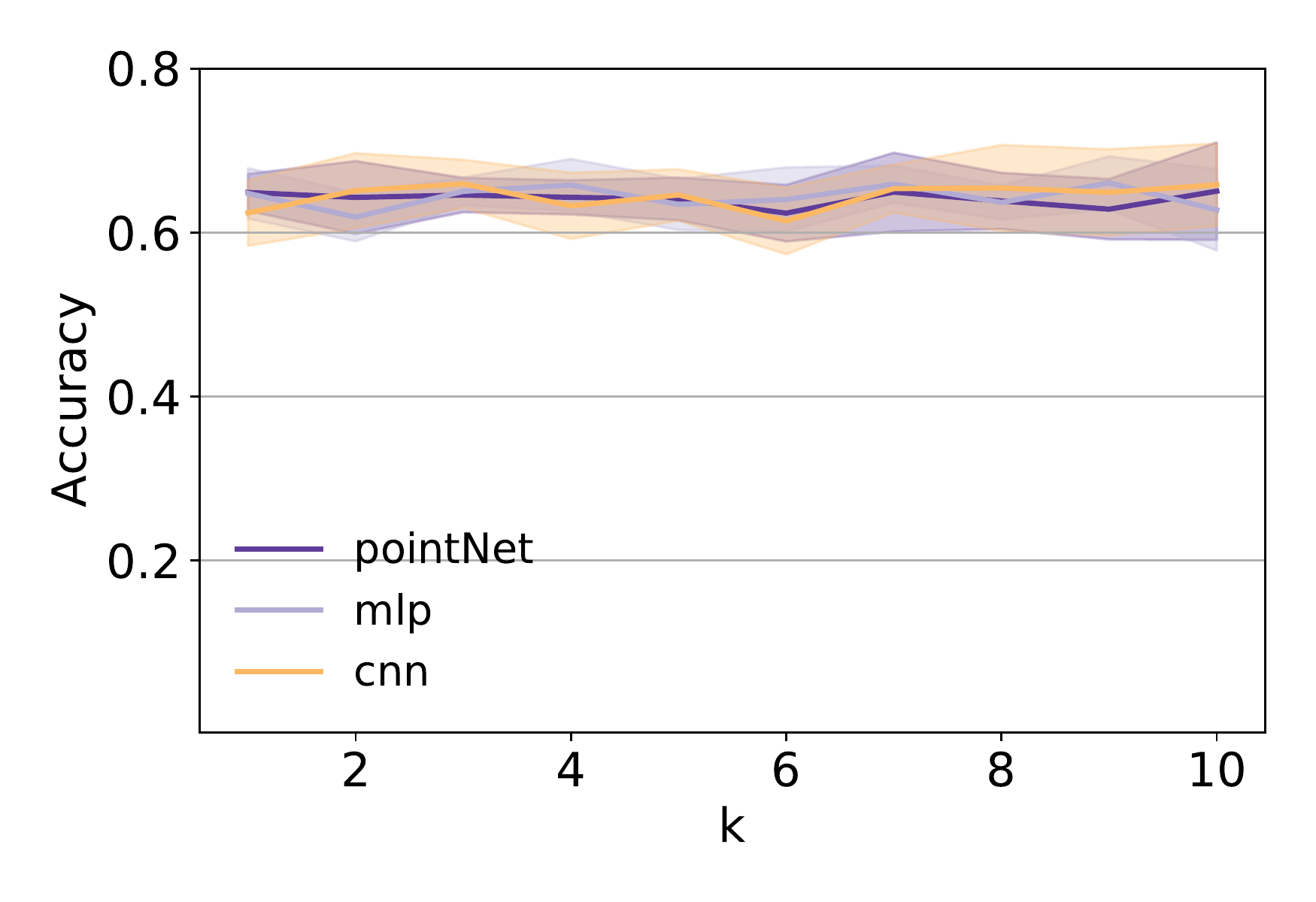}
    		\caption{Accuracy (Pubmed)}\label{fig:pubmed_acc}		
	\end{subfigure}
    \hfill
	\begin{subfigure}[t]{0.32\linewidth}
		\includegraphics[width=1\linewidth]{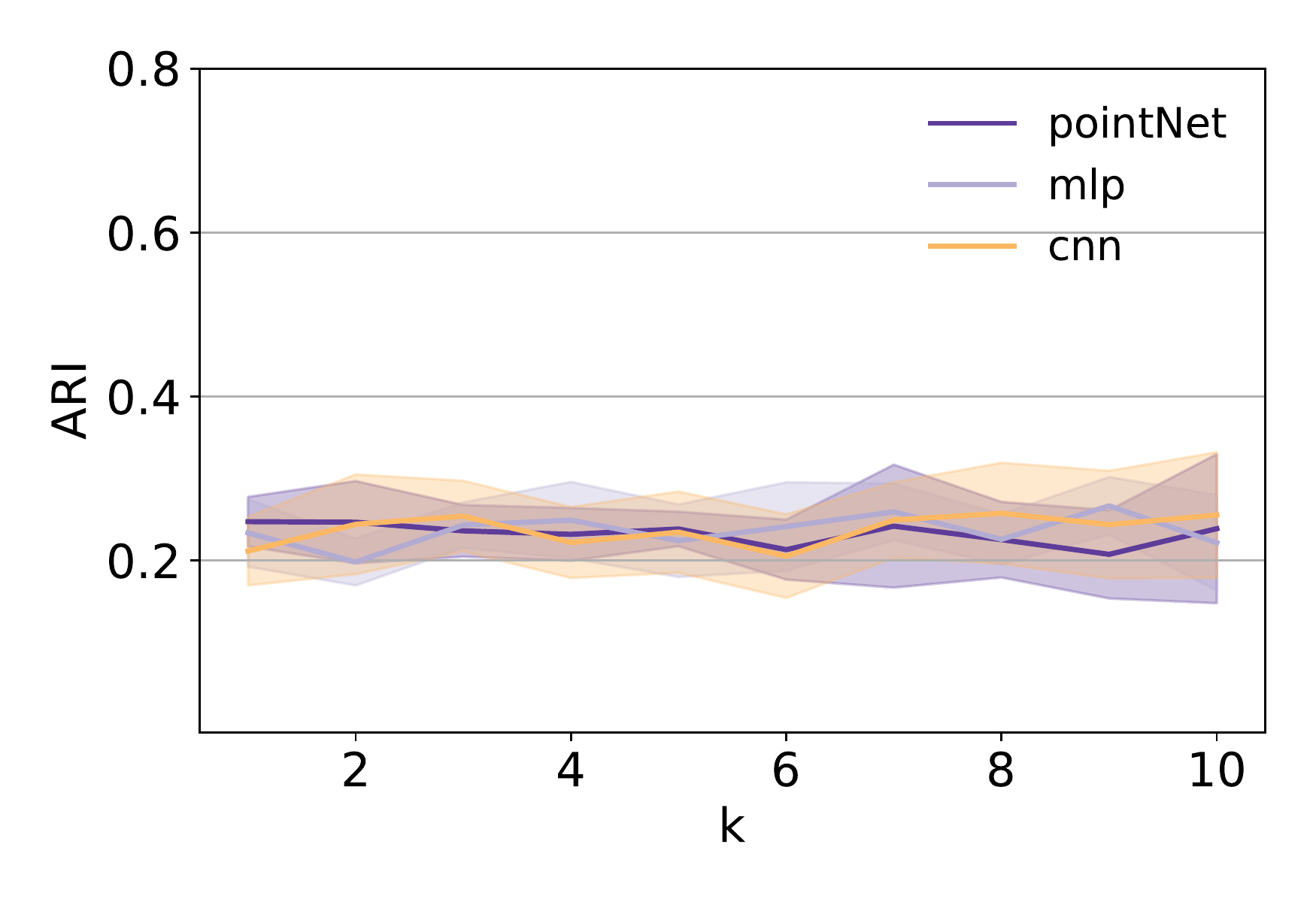}
		\caption{ARI (Pubmed)}\label{fig:pubmed_ari}	
	\end{subfigure}
    \hfill
	\begin{subfigure}[t]{0.32\linewidth}
		\includegraphics[width=1\linewidth]{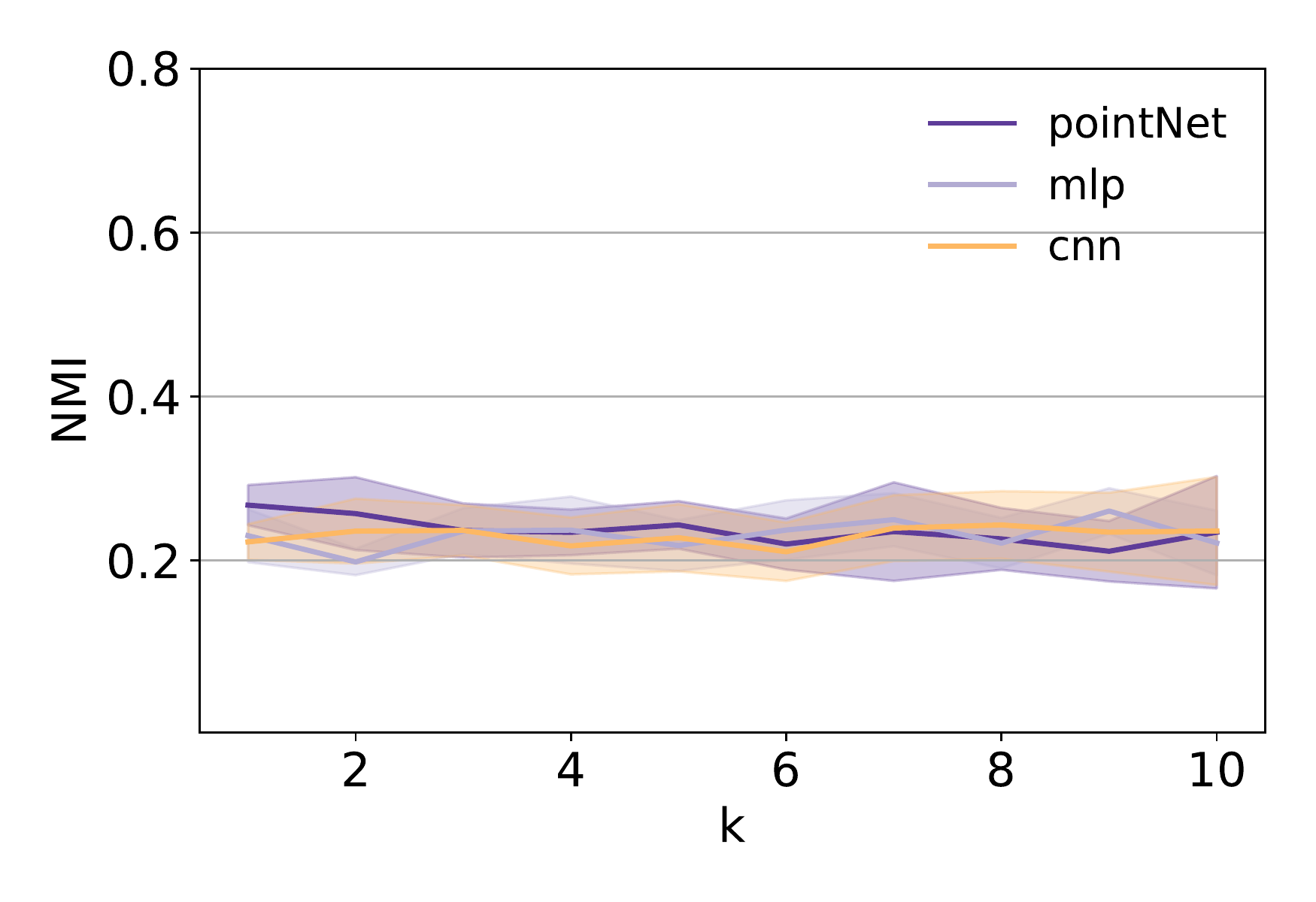}
		\caption{NMI (Pubmed)}\label{fig:pubmed_nmi}		
	\end{subfigure}
	\caption{Accuracy, ARI and NMI metrics for PointNetST, MLP and CNN-based networks solving the clustering task on Citeseer and Pubmed datasets. For all three measures, the mean value and standard deviation of 10 experiment runs are depicted.}	
	\label{fig:different_k_s_appendix}
\end{figure*}

Influence of parameter k - the convolution order - is also presented with respect to Citeseer and Pubmed in Figure~\ref{fig:different_k_s_appendix}. Citeseer presents similar behavior to Cora as it has similar graph structure with many node attributes being present. However, on Pubmed training behavior is slightly different. As shown, PointNetST does not perform constantly better than its CNN variant. This can be explained by the fact that node attributes are limited in Pubmed, while graph structure is dominant. Thus, the conventional neural networks can overfit on this structure and depict equally high performance compared to set equivariant ones.

\subsection{Training Convergence on Citeseer and Pubmed}
\label{sec:convergence_appendix}

\begin{figure*}[hbtp]
	\begin{subfigure}[t]{0.32\linewidth}
		\includegraphics[width=1\linewidth]{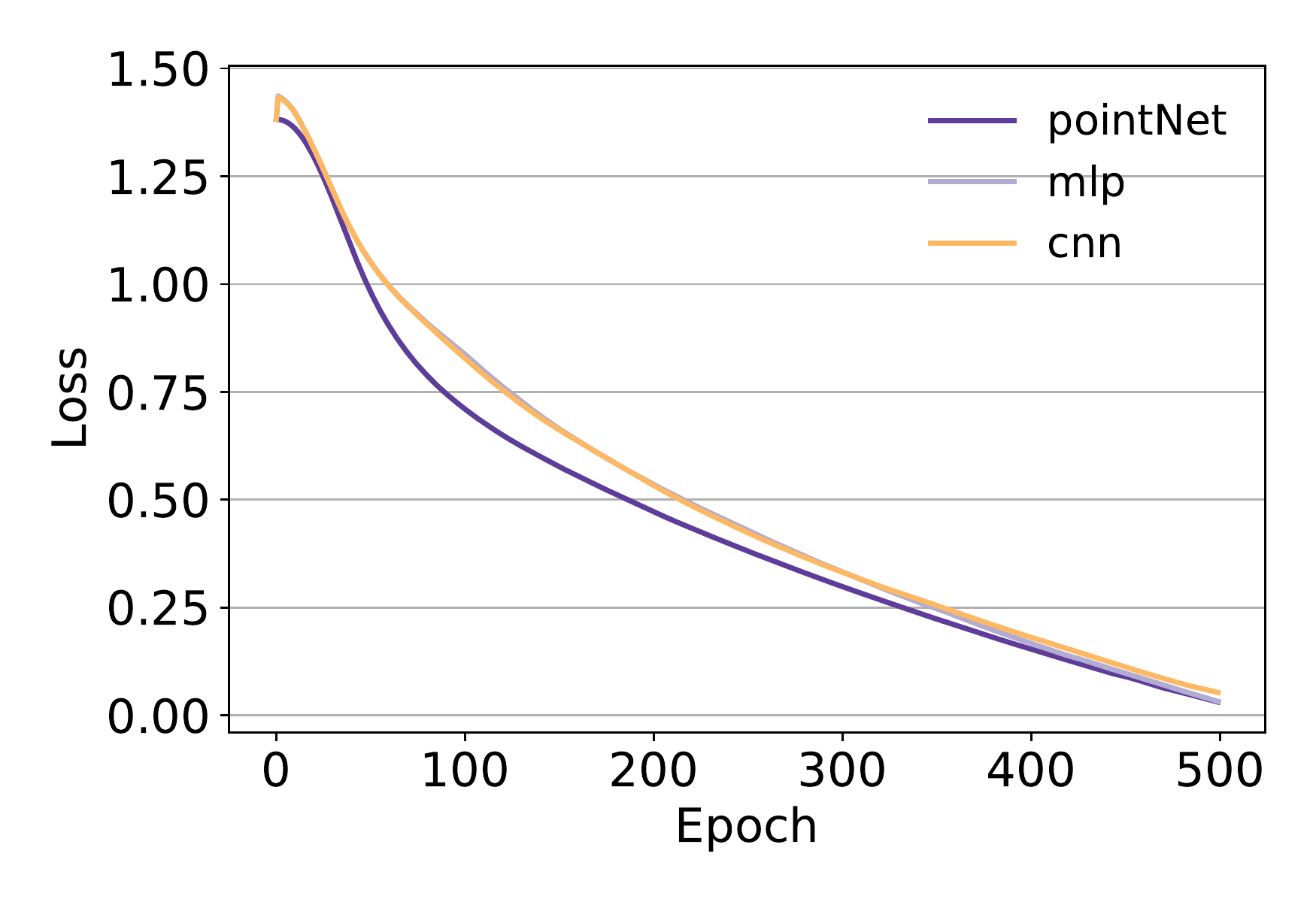}
    		\caption{Loss (Citeseer)}\label{fig:citeseer_loss}		
	\end{subfigure}
    \hfill
	\begin{subfigure}[t]{0.32\linewidth}
		\includegraphics[width=1\linewidth]{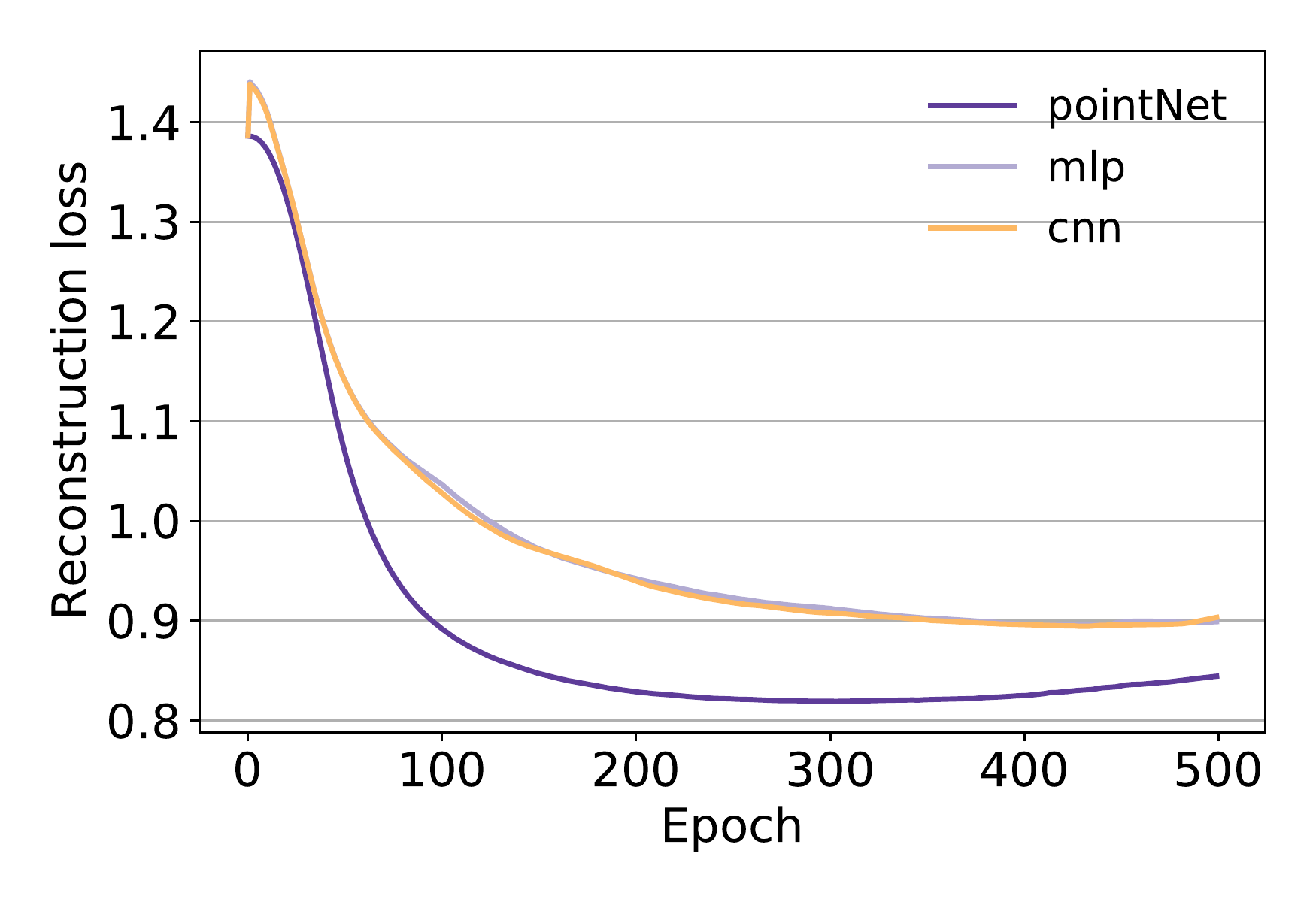}
		\caption{Reconstruction loss (Citeseer)}\label{fig:citeseer_r_loss}	
	\end{subfigure}
    \hfill
	\begin{subfigure}[t]{0.32\linewidth}
		\includegraphics[width=1\linewidth]{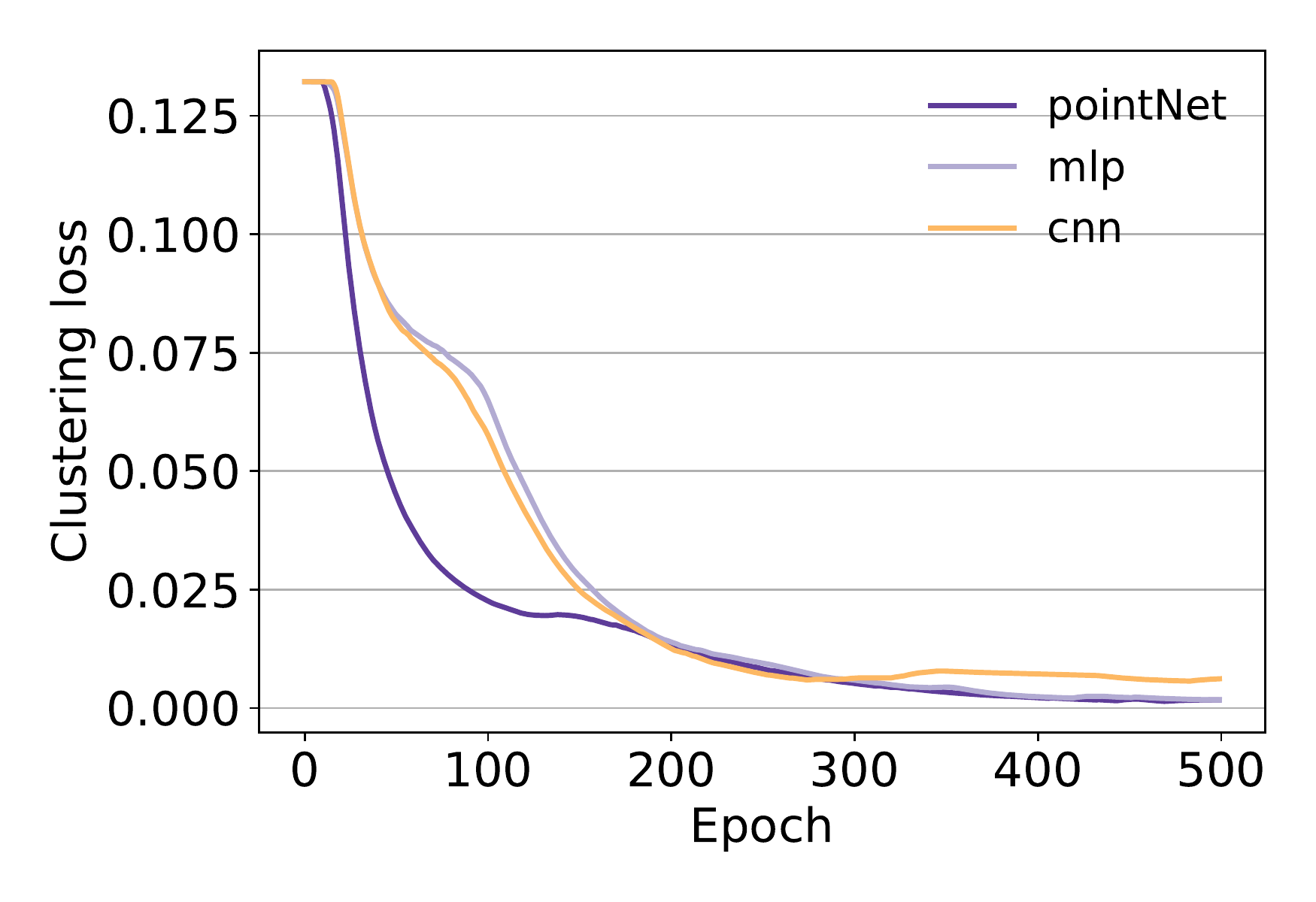}
		\caption{Clustering loss (Citeseer)}\label{fig:citeseer_c_loss}		
	\end{subfigure}
	\\
	\begin{subfigure}[t]{0.32\linewidth}
		\includegraphics[width=1\linewidth]{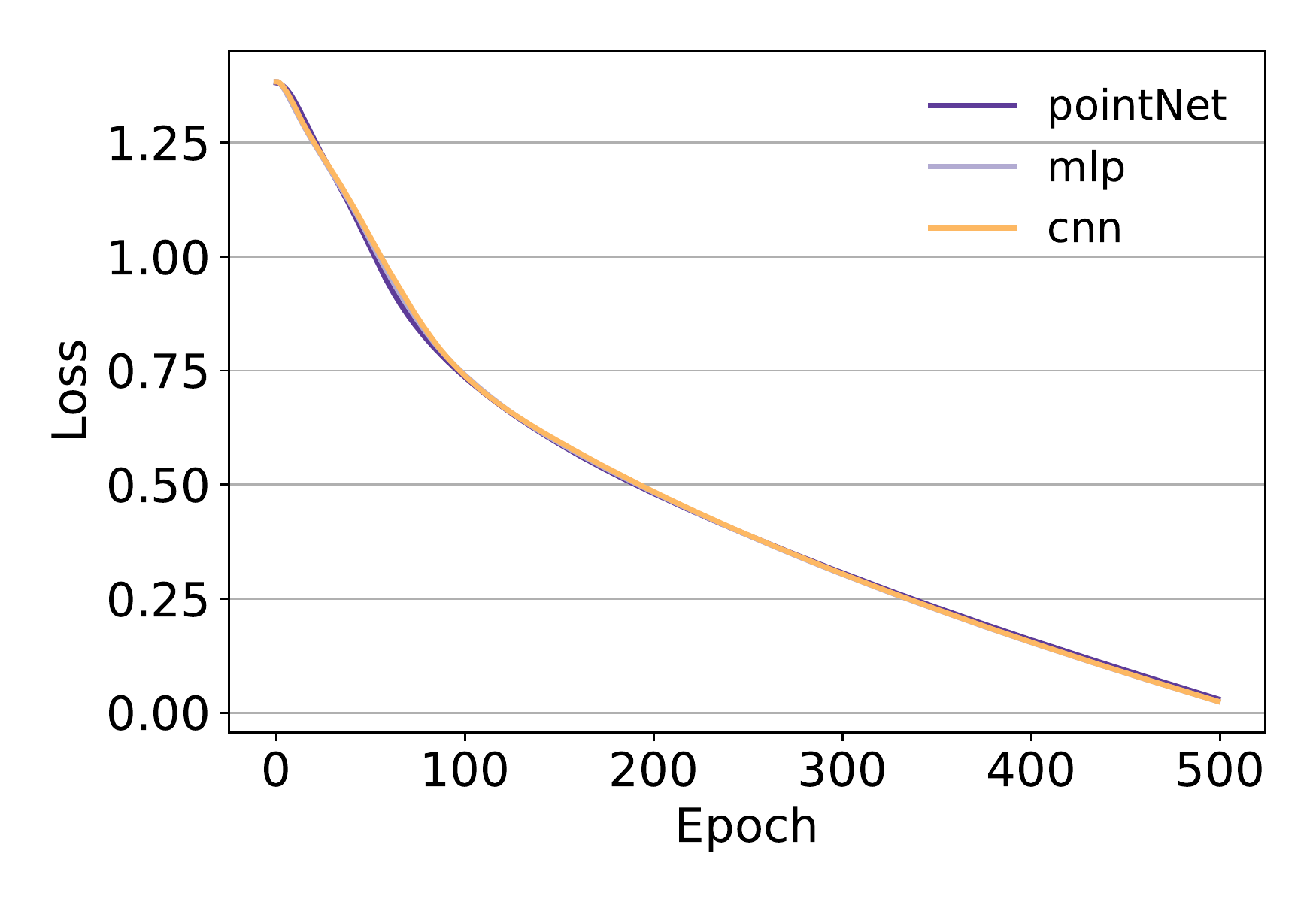}
    		\caption{Loss (Pubmed)}\label{fig:pubmed_loss}		
	\end{subfigure}
    \hfill
	\begin{subfigure}[t]{0.32\linewidth}
		\includegraphics[width=1\linewidth]{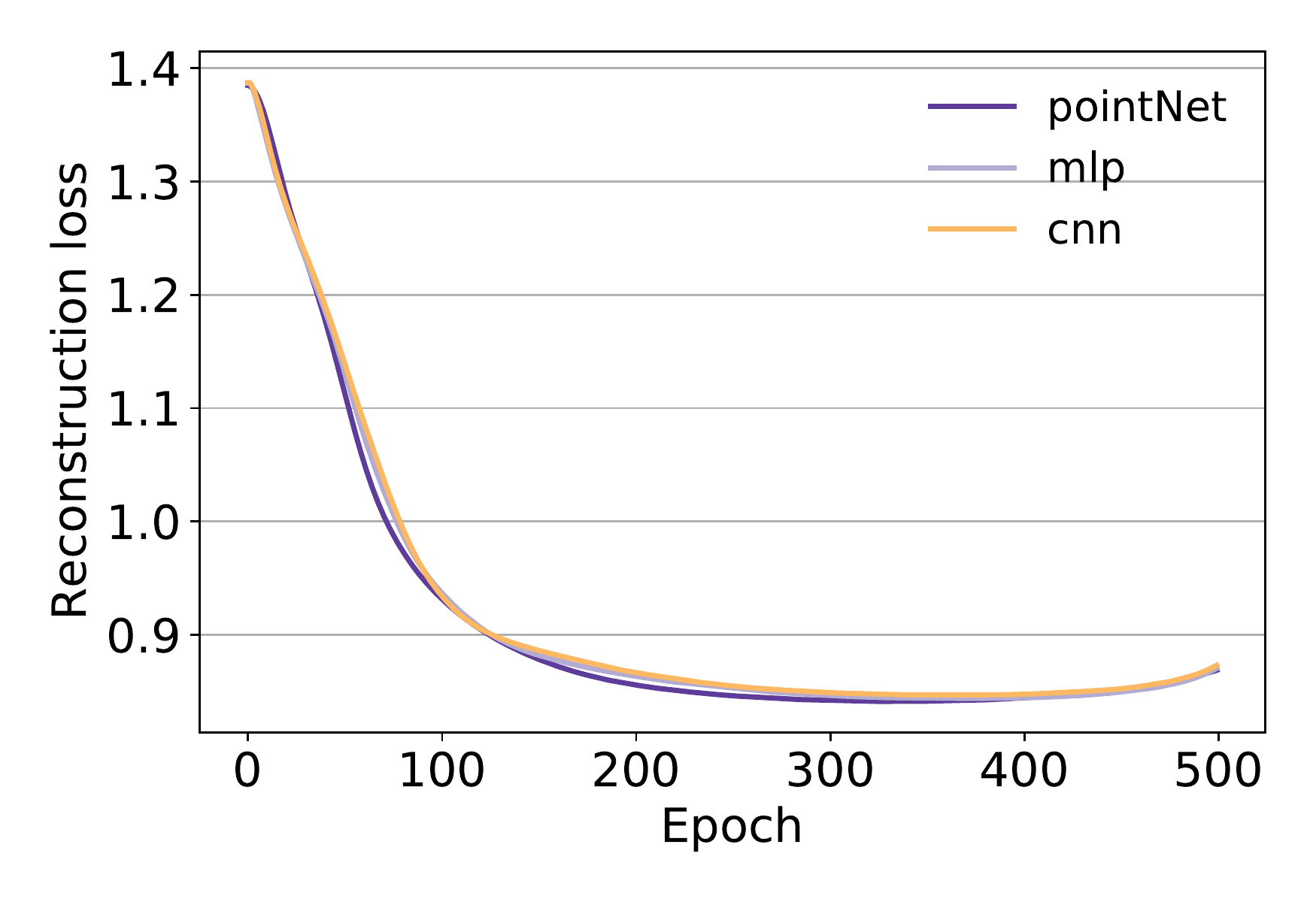}
		\caption{Reconstruction loss (Pubmed)}\label{fig:pubmed_r_loss}	
	\end{subfigure}
    \hfill
	\begin{subfigure}[t]{0.32\linewidth}
		\includegraphics[width=1\linewidth]{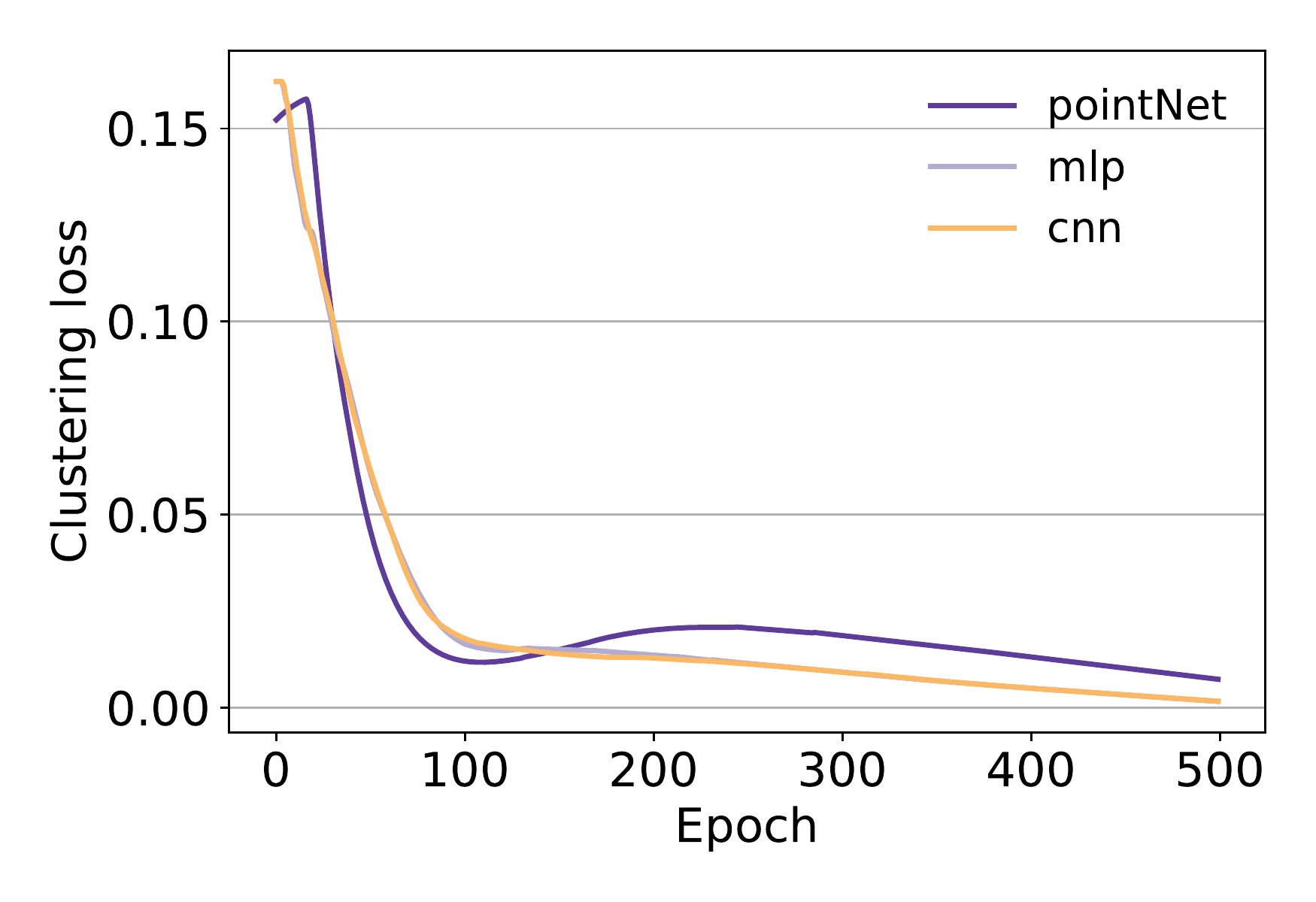}
		\caption{Clustering loss (Pubmed)}\label{fig:pubmed_c_loss}		
	\end{subfigure}
	\caption{Loss, reconstruction and clustering loss for PointNetST, MLP and CNN-based networks solving the clustering task on Citeseer and Pubmed datasets. For all three measures, the mean value and standard deviation of 10 experiment runs are depicted. PointNetST helps the model to converge faster than the MLP and CNN variants.}	
	\label{fig:convergence_appendix}
\end{figure*}

Extending the discussion on training convergence, Figure~\ref{fig:convergence_appendix} depicts the comparison between PointNetST encoder and MLP and CNN ones. Again, Citeseer behaves similar to Cora due to their similar structural and contextual information. Moreover, as already discussed Pubmed presents a different graph structure and less node information complicating set equivariance and restricting its performance. However, despite these complications PointNetST maintains its efficiency and accelerates the training convergence, although by a much smaller factor, nearly indistinguishable from the rest methods.

\subsection{Laplacian Smoothing and Graph Convolution}
\label{sec:smoothing-full}

The most important notion in the prevalent GNN-based embedding methods, such as GCN~\cite{kipf2016semi}, is that neighboring nodes should be similar and hence their features should be smoother - than that of irrelevant nodes - in the graph manifold. However, these methods capture deeper connections by stacking multiple layers, leading to deep architectures, which are known to overly smooth the node features~\cite{chen2020measuring}. To address this problem, the domain of graph signal processing considers $\mathbf{x} \in \mathbb{R}^n$ as a graph signal, where each one of the $n$ nodes is assigned a scalar. Then, the smoothness of signal $\mathbf{x}$ depicts the similarity between all of the graph nodes. To calculate smoothness, the \textit{Rayleigh quotient}~\cite{horn2012matrix} over the signal and the graph Laplacian $\mathbf{L}$ - essentially the normalized variance of $\mathbf{x}$ - is employed:
\begin{equation}
    R(\mathbf{L}, \mathbf{x}) = \frac{\mathbf{x}^T\mathbf{L}\mathbf{x}}{\mathbf{x}^T\mathbf{x}} = \frac{\sum_{(i,j) \in \mathcal{E}(x_i - x_j)^2}}{\sum_{i \in \mathcal{V}} x^2_i}
\end{equation}
Since neighboring nodes should be similar, a smoother signal is expected to have lower Rayleigh quotient. To find the relation between eigenvalues and Rayleigh quotient, one needs to calculate the eigendecomposition of the graph Laplacian, that is $ \mathbf{L} = \mathbf{U}\mathbf{\Lambda}\mathbf{U}^{-1} $ with $\mathbf{U} \in \mathbb{R}^{n \times n}$ being the matrix of eigenvectors and $\mathbf{\Lambda} = diag(\lambda_1, \lambda_2, \dots, \lambda_n)$ the diagonal matrix of eigenvalues. Then, the Rayleigh quotient of the eigenvector $\mathbf{u_i}$ is:
\begin{equation}
\label{eq:rayleigh_eig}
    R(\mathbf{L}, \mathbf{u_i}) = \frac{\mathbf{u_i}^T\mathbf{L}\mathbf{u_i}}{\mathbf{u_i}^T\mathbf{u_i}} = \lambda_i
\end{equation}

It can be seen that the lower Rayleigh quotients - and by extension the smoother eigenvectors - are correlated with low eigenvalues, meaning low frequencies. To employ these observations to every signal $\mathbf{x}$, the decomposition of x on the basis of $\mathbf{L}$ is considered:
\begin{equation}
\label{eq:x_decomp}
    \mathbf{x} = \mathbf{U} \mathbf{p} = \sum_{i=1}^n p_i\mathbf{u_i}
\end{equation}

Consequently, as smooth signals are associated with smooth eigenvectors and low eigenvalues according to Eq.~\ref{eq:rayleigh_eig}, the used filter should cancel high frequencies and preserve the low ones. Laplacian smoothing filters are selected for this purpose, as they combine high performance with low computational cost~\cite{taubin1995signal}.

\noindent \textbf{Laplacian Smoothing Filter:} Here, we consider the generalized Laplacian Smoothing filter as defined in~\cite{taubin1995signal}
\begin{equation}
\label{eq:gl_app}
    \mathbf{H} = \mathbf{I} - k\mathbf{L} 
\end{equation}

where $k \in \mathbb{R}$, I is the identity matrix and $\mathbf{H}$ is the filter matrix. Using Eq.~\ref{eq:gl_app}, the filtered signal is:
\begin{align*}
\label{eq:smoothed_signal}
    \mathbf{\Bar{x}} & = \mathbf{H}\mathbf{x} = \mathbf{U}(\mathbf{I} - k\mathbf{\Lambda})\mathbf{U}^{-1}\mathbf{U}\mathbf{p} \\
    & = \sum_{i=1}^n (1-k\lambda_i)p_i\mathbf{u_i}
\end{align*}

What this suggests is that for $\mathbf{H}$ to be low-pass, $1-k\lambda$ should always decline. It has been found that the optimal value of $k$ is $1/\lambda_{max}$, with $\lambda_{max}$ denoting the largest eigenvalue~\cite{cui2020adaptive}. 

Having defined the filter, one can introduce k-order smoothing - and thus graph convolution - by stacking k filters together. Ultimately, the overall smoothed feature matrix is
\begin{equation}
    \mathbf{\Bar{X}} = \mathbf{H^k}\mathbf{X}
\end{equation}

\subsection{System Specifications}
\label{sec:system_specs_appendix}

All of the experiments were conducted using a computing grid with an Intel Xeon E5-2630 v4 CPU, 32Gb RAM and an Nvidia Tesla P100 GPU. Also, to speed up some computations a personal computer with an Intel(R) Core(TM) i7-6700K CPU @ 4.00GHz, 32Gb RAM and an NVidia GeForce GTX 1070 GPU was also employed for specific experiments alongside the ones running on the grid.